%% file: main.tex
\documentclass[10pt,twocolumn,letterpaper]{article}

\usepackage{cvpr}              % To produce the CAMERA-READY version
\input{preamble}

\definecolor{cvprblue}{rgb}{0.21,0.49,0.74}
\usepackage[pagebackref,breaklinks,colorlinks,allcolors=cvprblue]{hyperref}

\def\paperID{} % *** Enter the Paper ID here
\def\confName{CVPR}
\def\confYear{2025}

\title{Camera-Only 3D Panoptic Scene Completion for Autonomous Driving through Differentiable Object Shapes}

\author{Nicola Marinello$^{1}$ \and Simen Cassiman$^{1}$ \and Jonas Heylen$^{2}$ \and Marc Proesmans$^{1,2}$ \and Luc Van Gool$^{1,2,3,4}$\\
    $^1$ KU Leuven \quad $^2$ TRACE vzw \quad $^3$ ETH Zürich \quad $^4$ INSAIT \\
    \small{\texttt{\{nicola.marinello,simen.cassiman,marc.proesmans,luc.vangool\}@esat.kuleuven.be}}\\
    \small{\texttt{jonas.heylen@trace.vision}}
}

\newcommand{\ourmodelname}{OffsetOcc} % usage: \ourmodelname{}
\newcommand{\paragrax}[1]{{\bf #1}\;}

\newcommand{\indic}[1]{\mathds{1}_{\{#1\}}}

\begin{document}

\begin{acronym}[ICANN]
    \acro{iou}[\textsc{i}o\textsc{u}]{intersection over union} %\ac{iou}
    \acro{miou}[m\textsc{i}o\textsc{u}]{mean intersection over union}
    %% Define a custom plural form of an acronym   
    \acrodefplural{cnn}[\textsc{CNN}s]{Convolutional Neural Networks} % \aclp{cnn}
\end{acronym}

\maketitle
\input{sec/0_abstract}
\input{sec/1_intro}

\input{sec/2_relwork}
\input{sec/3_method}

\input{sec/4_experiment}
\input{sec/5_conclusion}
{
    \small
    \bibliographystyle{ieeenat_fullname}
    \bibliography{main}
}

\clearpage
\appendix
\input{sec/X_suppl}

\end{document}

%% file: preamble.tex
\usepackage{lipsum}
\usepackage{dsfont}
\usepackage{makecell}
\usepackage{subcaption}
\usepackage{graphicx}
\usepackage{capt-of}  % <---
\usepackage{cuted}    % <===
\usepackage{lineno} % not sure why it is needed?
\usepackage[nolist,nohyperlinks]{acronym}
\usepackage[accsupp]{axessibility}  % Improves PDF readability for those with disabilities.

\usepackage{pifont}
\newcommand{\cmark}{\ding{51}}%
\newcommand{\xmark}{\ding{55}}%

\usepackage{dirtytalk}
\MakeRobust{\say}
\input{misc/occupancy_colors}

%% file: misc/occupancy_colors.tex
\definecolor{nbarrier}{RGB}{255, 140, 0}
\definecolor{nbicycle}{RGB}{255, 61, 99}
\definecolor{nbus}{RGB}{255, 247, 0}
\definecolor{ncar}{RGB}{255, 158, 0}
\definecolor{nconstruct}{RGB}{233, 150, 70}
\definecolor{nmotor}{RGB}{112, 128, 144}
\definecolor{npedestrian}{RGB}{0, 0, 230}
\definecolor{ntraffic}{RGB}{233, 150, 70}
\definecolor{ntrailer}{RGB}{0, 0, 0}
\definecolor{ntruck}{RGB}{47, 79, 79}
\definecolor{ndriveable}{RGB}{0, 207, 191}
\definecolor{nother}{RGB}{255, 0, 0}
\definecolor{nsidewalk}{RGB}{75, 0, 75}
\definecolor{nterrain}{RGB}{165, 42, 42}
\definecolor{nmanmade}{RGB}{128, 128, 128}
\definecolor{nvegetation}{RGB}{0, 175, 0}
\definecolor{nothers}{RGB}{0, 0, 0}

%% file: sec/0_abstract.tex
\begin{abstract}
Autonomous vehicles need a complete map of their surroundings to plan and act. This has sparked research into the tasks of 3\textsc{d} occupancy prediction, 3\textsc{d} scene completion, and 3\textsc{d} panoptic scene completion, which predict a dense map of the ego vehicle\textquotesingle s surroundings as a voxel grid. Scene completion extends occupancy prediction by predicting occluded regions of the voxel grid, and panoptic scene completion further extends this task by also distinguishing object instances within the same class; both aspects are crucial for path planning and decision-making. However, 3\textsc{d} panoptic scene completion is currently underexplored. This work introduces a novel framework for 3\textsc{d} panoptic scene completion that extends existing 3\textsc{d} semantic scene completion models. We propose an \emph{Object Module} and \emph{Panoptic Module} that can easily be integrated with 3\textsc{d} occupancy and scene completion methods presented in the literature. Our approach leverages the available annotations in occupancy benchmarks, allowing individual object shapes to be learned as a differentiable problem. The code is available at \url{https://github.com/nicolamarinello/OffsetOcc}.
\end{abstract}

%% file: sec/1_intro.tex
\section{Introduction}
\label{sec:intro}

Computer Vision (\textsc{cv}) \cite{Szeliski2022} is, in large part, concerned with extracting conceptual information from sensory data. The field has seen significant advancements, largely driven by improvements in Deep Learning \cite{Lecun2015}, allowing it to tackle increasingly difficult tasks. For the better part of the last decade, progress was mainly viewed in light of benchmarks, providing handcrafted, narrow, and sometimes artificial metrics. However, recent advances have allowed methods to go beyond these isolated experiments to their incorporation into various applications, showcasing their potential in the real world.

\input{images/teaser_image}

One of the most useful applications may well be enabling autonomous agents, analyzing sensory data to build up an accurate representation of their surroundings to guide their decision-making. This has motivated research into 3\textsc{d} object detection \cite{Mao2023}, which deals with localizing objects in 3\textsc{d} space, and the generation of birds-eye-view (\textsc{bev}) maps \cite{Li2022}, which extract a 2\textsc{d} top view of the entire scene. However, while performance has improved greatly, these methods leave out important information for subsequent operational modules such as prediction and planning. Detection models output bounding boxes, which only provide coarse object shape estimates, ignoring uncountable classes (referred to as \textit{stuff}, \eg road) \cite{Tian2023}. In contrast, \textsc{bev} maps handle stuff and can provide more accurate shapes on the ground plane, but they lose important height information, which is hard to recover and could impact downstream tasks.

These shortcomings lead to the introduction of Semantic Occupancy Prediction \cite{Xu2023, Zhang2024}, the task of labeling each visible point in space. In the case of autonomous driving, initial methods labeled \textsc{lidar} point clouds, but these are usually very sparse and leave out a lot of general knowledge about scene structure that is, again, very important to solve the actual end task of autonomy. Therefore, more recent work handles voxel occupancy prediction \cite{Tian2023, Tong2023, Wang2023}, tasked with completing a full 3\textsc{d} grid of the scene (\ie semantic labeling of 3\textsc{d} or volumetric pixels, known as \emph{voxels}). The latest research moves to camera-only settings \cite{Cao2022}, improving the learned features \cite{Li2023, Zhang2023, Zheng2024} and their efficiency \cite{Huang2023, Yu2023, Sze2024, Ma2024, He2024, Zhao2024, Li2024, Wang2024, Huang2024}.

It is important to note that occupancy prediction aims to reconstruct only the directly detectable voxels (\ie detected in the current frame by the camera and/or \textsc{lidar}), which should be distinguished from Semantic Scene Completion \cite{Xu2023, Tian2023, Ma2024}, which intends to also reconstruct regions that are currently occluded by other portions of the scene. This is a very important distinction with big implications for real-world applications. Most models evaluate the reconstruction performance by using a \say{camera mask}, which indicates which voxels are directly detectable in the current frame, since these regions are determined with relative certainty and can be considered ground truth, in contrast to undetected regions. However, while knowing how well a method handles directly visible areas is certainly important, it is an incomplete task, since this artificial mask that knows what has been detected and what has not is not available in a real-world situation. Furthermore, models trained with this mask exploit hidden regions to improve their accuracy \cite{Liu2023}. This can give a false sense of performance by discarding predictions in occluded regions (which might be very bad) that are not distinguishable during inference. Moreover, it is clear that humans can infer information about some of the occluded areas, especially about partially occluded objects, indicating that \textsc{cv} methods must share this ability to level the playing field for reasoning capabilities. Therefore, it is important to develop and evaluate models for these hidden regions as well; visible portions must be reconstructed accurately, and occluded areas should contain reasonable predictions.

Furthermore, the planning and decision-making of an autonomous agent also depend on the actions of other agents in the scene. Therefore, it is important to distinguish each instance within the same class. This has led to models that can produce panoptic occupancy and scene completion outputs \cite{Wang2024PanoOcc,Yu2024}. However, this topic is underexplored, and many possibilities to distinguish instances remain to be tested or developed.

This work presents a 3\textsc{d} panoptic scene completion framework for autonomous driving using only camera inputs. It achieves this by using a separate object decoder through which shapes can be learned differentiably and per class. Conventional methods that uplift 2\textsc{d} features to 3\textsc{d} and leverage self-attention mechanisms effectively infer overall object occupancy. However, approaches that distinguish individual object instances within the scene remain underexplored. Notably, the proposed method can be seamlessly integrated with more advanced occupancy decoders and enhanced with temporal reasoning, as explored in recent works.

Our contributions are summarized as follows:

\begin{itemize}
    \item We present a novel framework that models individual object occupancy as a differentiable problem.
    \item We present an \emph{Object Module} and \emph{Panoptic Module} that can extend existing 3\textsc{d} semantic scene completion methods, enabling panoptic 3\textsc{d} scene completion.
    \item We release our implementation to promote further research.
\end{itemize}

%% file: images/teaser_image.tex
\begin{figure}[h]
    \centering
    \includegraphics[width=\columnwidth]{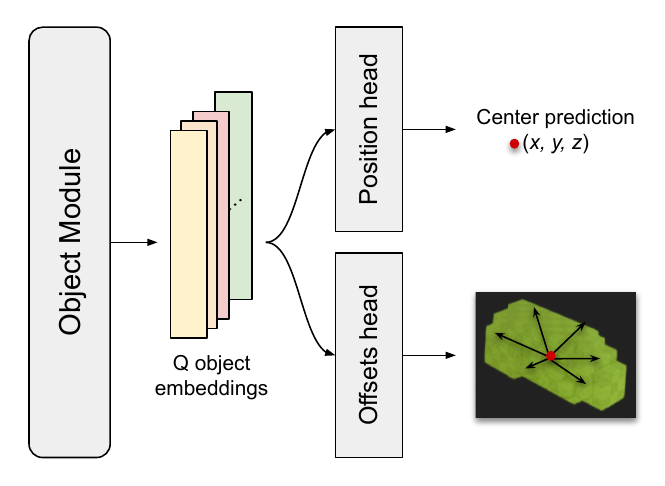}
    \caption{\textbf{\ourmodelname{}.} Our method \ourmodelname{} models object occupancy as a set prediction task, where objects are represented by their 3\textsc{d} center and a set of offsets that describe their shape. The set of offsets is learned from the dataset annotations. Classification head omitted for conciseness.}
    \label{fig:teaser}
\end{figure}

%% file: sec/2_relwork.tex
\section{Related work}
\label{sec:related}

Many works in this field primarily focus on semantic occupancy prediction, trying to improve the learned features, increase model efficiency, or tackle the complex label annotation problem. Fewer works investigate the scene completion objective, which is closely related and can often benefit from improvements to the former. Lastly, very few, and only recently, have proposed panoptic scene completion methods to distinguish the different object instances within the same class.

\noindent{\paragrax{Semantic Occupancy Prediction.}} To learn better features, Li \etal \cite{Li2023} propose using additional losses such as depth and 2\textsc{d} segmentation, also explored by Zheng \etal \cite{Zheng2024} in the form of a semantic loss and additional temporal information from past frames, a practice that is becoming increasingly popular. One branch of works tackling efficiency handles it by way of a lighter representation, such as keeping the \textsc{bev} as proposed by Yu \etal \cite{Yu2023}, a hybrid model with \textsc{bev} and \textsc{bev} slices put forth by Zhang \etal \cite{Zhang2023}, or \textsc{bev} and voxels by He \etal \cite{He2024}, extending the 2\textsc{d} representation to triplane view (\textsc{tpv}) as presented by Huang \etal \cite{Huang2023}, or using a low-rank approximation with vertical vectors and horizontal matrices, per Zhao \etal \cite{Zhao2024}. Tian \etal \cite{Tian2023} propose using a full voxel grid, but starting with a coarser version that gets refined until the output dimension. Others reduce computational cost by choosing different operations, like Sze \etal \cite{Sze2024}, who use sparse convolutions; Wang \etal \cite{Wang2024}, who use a linear complexity attention with Receptance Weighted Key Value operations \cite{Peng2023}; Li \etal \cite{Li2024}, who exchange transformers for Mamba \cite{Gu2023}; or Huang \etal \cite{Huang2024}, who use Gaussians to model regions of interest, only allocating compute where it is required, with the additional benefit that the decoding is object-centric. Wang \etal \cite{Wang2024Reli} explore reliability by incorporating uncertainty learning, an aspect that is also handled by Wang \etal \cite{Wang2024Gen} using diffusion models, which naturally incorporate a coarse-to-fine iterative refinement. Lastly, Pan \etal \cite{Pan2023} try to avoid dense labels by using \textsc{nerf}s and 2\textsc{d} labeling, which Pan \etal \cite{Pan2024} follow up by adding a teacher-student pseudo-labeling setup. Huang \etal \cite{Huang2024Self} even go a step further, using self-supervised learning by predicting the frames before and after.

While there has been great progress thanks to these works, it is only shown for visible areas, ignoring the many occluded regions that contain relevant information for modules further in the pipeline of autonomous vehicles. Moreover, by training these methods with the mask, they learn unrealistic object shapes, which can actually be more different from each other and, therefore, potentially make learning general concepts more difficult.

\noindent{\paragrax{Semantic Scene Completion.}} In contrast to the previous paragraph, scene completion methods aim to reasonably reconstruct the entire scene, including occluded areas. Cao \etal \cite{Cao2022} use a monocular system to show that \textsc{rgb} camera-only setups can also provide reasonable outputs, even without additional depth information. Li \etal \cite{Li2023VoxFormer} achieve this by explicitly handling occupancy and completion separately, first reconstructing the visible voxels and then using a masked autoencoder to complete the occluded areas. Wei \etal \cite{Wei2023} employ Poisson Reconstruction \cite{Kazhdan2006} to fill in holes and use it to generate denser ground truth. Due to the high sparsity, Ma \etal \cite{Ma2024} propose to work in a strongly downsampled representation, where all operations can be computed more efficiently, after which it is upsampled and classes are predicted per mask. Lastly, sparsity is also exploited by Liu \etal \cite{Liu2023}, by only using sparse operations until the very end. To increase performance, they also add temporal information. Additionally, they propose RayIoU to deal with the inaptitude of \ac{miou}, the dominant metric in the field, when handling depth errors.

Despite extracting more information by filling in invisible parts, these methods rely on the models to learn the relationships between classes and their plausible shapes that are required to complete occluded areas. If a car is only partially visible, the model must learn that voxels in the occluded region should be related to it, which is a very complex learning problem, especially when labels are imperfect in these regions.

\noindent{\paragrax{Panoptic Scene Completion.}} Wang \etal \cite{Wang2024PanoOcc} combine 3\textsc{d} detection and segmentation into one framework, using sparsity techniques for efficiency and temporal reasoning for improved completion. By incorporating detection, they can separate the instances within each class. Yu \etal \cite{Yu2024} build on Flash-Occ for its efficiency, first predicting occupancy and then combining the related voxels into instances.

However, instance voxels are decoded individually, meaning that models must learn what it means to be part of an object, and that objects have only a single class. Within the voxel decoding modules, there is no notion of an entity to which a voxel belongs; they are decoded bottom-up, and then the instance is estimated. This discards valuable top-down, object-centric information and, therefore, seems suboptimal.

%% file: sec/3_method.tex
\section{Methodology}
\label{sec:method}

\subsection{Problem setup}

3\textsc{d} semantic scene completion is the task of reconstructing the complete surrounding environment of an ego vehicle, including occluded regions (within reason, \ie it cannot be reasonably expected that objects which have never been seen, such as those behind a building corner, are also predicted). This work focuses on the case where only camera inputs are available (\ie no \textsc{lidar} inputs are used). The environment is represented as a voxel grid, with each voxel receiving a semantic label. Given the observations from \(N\) cameras at a specific timestamp \(\mathbf{I}_t = \left\{ I_t^1, I_t^2, \dots, I_t^N \right\}\), the goal is to generate a grid \(\mathbf{\hat{Y}}_t \in \{c_0, c_1, \dots, c_M\}^{L \times W \times H}\), where \(c_0\) denotes the empty class to indicate unoccupied space, \(\{c_1, \dots, c_M\}\) denote the semantic classes, while \(L\), \(W\), and \(H\) denote the length, width, and height of the grid, respectively. The task can be extended to panoptic scene completion, where all voxels occupied by the same object instance are assigned the same object \textsc{id}, which is unique across all objects.

\subsection{Object occupancy as a set prediction}

We propose a novel framework \emph{\ourmodelname{}} for the 3\textsc{d} panoptic scene completion task that models individual objects and their shapes as a differentiable set prediction problem. Inspired by \textsc{detr} \cite{carion2020end}, we model each object's occupied voxels as a set of 3\textsc{d} vectors relative to the center of the object, as shown in \cref{fig:teaser}. Each of these vectors is referred to as an \emph{offset}, which aims to point towards the center of one of the voxels occupied by the object. This modeling allows us to effectively represent the occupancy of the object within the 3\textsc{d} grid. Such a representation is more efficient than computing a full grid mask of each object, which becomes computationally prohibitive. This approach facilitates a structured and precise representation of object shapes, supporting downstream tasks that require accurate object estimation.

\subsection{Architecture}
\label{sec:architecture}

\input{images/architecture}

In this section, we detail the architecture of our model. We present an \emph{Object Module} and \emph{Panoptic Module}, which can be used on top of a generic baseline model. The different components of this baseline can be instantiated with various methods proposed in the literature.

As shown in \cref{fig:architecture}, the model takes multi-view images as input, from which features are extracted via a backbone. Subsequently, an \emph{Encoder} uses these camera features to generate 3\textsc{d} representations in a (downsampled) voxel grid, which we will refer to as voxel embeddings. These then pass through the \emph{Decoder} to generate the baseline voxel occupancy predictions. Our proposed \emph{Object Module} also takes these voxel embeddings as input and provides instance-level occupancy predictions. Lastly, the baseline voxel occupancy prediction and the instance-level predictions are passed to our proposed \emph{Panoptic Module}, which generates the final panoptic voxel occupancy. The structure and responsibility of each module are further detailed in the paragraphs below. Since the proposed modules are meant to work with generic baselines, we focus on the structure of our proposed method; specific implementations of the baseline modules are detailed in \cref{sec:experiments-implementation}.

\noindent{\paragrax{Backbone.}} A multi-view image feature extractor of choice. A common setup in the literature is the combination of a \textsc{cnn} backbone with an \textsc{fpn} on top for multi-level feature extraction.

\noindent{\paragrax{Encoder.}} The encoder lifts 2\textsc{d} image features to a 3\textsc{d} representation, providing a per-voxel feature vector. Optionally, the voxel grid at this stage can be predicted as a downsampled version of the final prediction to reduce the computational burden.

\noindent{\paragrax{Decoder.}} The decoder takes in the voxel embeddings and computes the final label predictions for each location in the grid. This prediction is at full resolution; thus, if the intermediate voxel grid was downsampled, the decoder must also take care of upsampling.

\noindent{\paragrax{Object Module.}} Inspired by \textsc{detr}’s detection approach \cite{carion2020end}, we extend its capabilities to predict object shapes in the form of voxel occupancy, in addition to their class and location. Our method utilizes a stack of \(M\) transformer layers composed of 3\textsc{d} deformable cross-attention \cite{Tong2023, zhu2020deformable} and self-attention layers. Starting with \(Q\) learnable object queries, the model produces \(Q\) object embeddings using the cross-attention layers to extract contextual information from the voxel feature grid, while self-attention layers allow the exchange of information between object embeddings. After this stack, the embeddings are further enhanced by a feed-forward network (\textsc{ffn}), and passed to a class head, a position head, and an offsets head, which predict the object’s class, the 3\textsc{d} object center location, and shape occupancy, respectively. Specifically, for each object \(i\), we predict the label \(\hat{y}^{i} \in \{c_0, c_1, \dots, c_M\}\), the 3\textsc{d} location \(\mathbf{\hat{c}}^{i} = (x, y, z)\), and a set of \(K\) offsets relative to the center of the object. The set of offsets for an object is defined as \(\mathbf{\hat{o}^{i}} = \{\mathbf{\hat{o}}_{1}^{i}, \dots, \mathbf{\hat{o}}_{K}^{i}\}\), where each \(\mathbf{\hat{o}}_{k}^{i} = (o_{k,x}^{i}, o_{k,y}^{i}, o_{k,z}^{i})\) is a 3\textsc{d} vector. For each offset, the linear layer also predicts an occupancy score \(\hat{s}_{k}^{i}\). The occupancy score spans the range \([0, 1]\). Formally, we describe an object occupancy as follows. First, given the object center and the set of offsets, we define the object point cloud as the set 
\begin{equation}
      \hat{V}^{i} = \{\mathbf{\hat{v}}_{k}^{i} = \mathbf{\hat{c}}^{i} + \mathbf{\hat{o}}_{k}^{i},  \forall k \in \{ 1, 2, \dots, K \} \mid \hat{s}_{k}^{i} \geqslant 0.5\}
\end{equation}Subsequently, an occupancy mask is obtained by voxelizing the resulting point cloud. The number of offsets \(K\) is a parameter and must be sufficiently large to describe objects of any size. The model is trained to assign a score \(\geq 0.5\) to a subset of offsets \(K' \leqslant K\). The directions of offsets with a score \(\geq 0.5\) aim to align with the centers of the grid voxels occupied by the object. Offsets with a score below \(0.5\) are discarded, as they are predicted to be unnecessary to describe the object occupancy.
Within this framework, the object center localizes each object within the grid, while the relative offsets describe its occupancy shape. This framework allows us to model the object's occupancy within the grid as a continuous and differentiable problem. Objects classified as background are discarded. By extending the \textsc{detr} approach, our model inherently achieves panoptic capabilities, as each object's occupancy is decoded independently from a separate query.

\noindent{\paragrax{Panoptic Module.}} This module is responsible for merging the occupancy predictions from the baseline and the \emph{Object Module} to generate panoptic occupancy predictions. We adopt a simple, parameter-free approach, where each voxel classified as an object by the baseline decoder is assigned an instance \textsc{id} based on the \emph{Object Module} prediction. To smooth out the prediction and correct small localization errors, we use a voting scheme involving the voxels within radius \(r\) of the reference location. The most voted \textsc{id} (\ie the \textsc{id} that is assigned most often to the sampled set of nearby voxels) is assigned to the reference voxel. If no meaningful \textsc{id} is sampled, then the empty class is assigned.

\subsection{Training methodology}
\label{sec:train_methodology}

In this section, we present our model's training methodology, which uses two stages. In the first stage, the baseline model is trained using the semantic labels of the occupancy grid according to \(\mathcal{L_{\textsc{ssc}}}\). In the second stage, the \emph{Object Module} and \emph{Panoptic Module} are added and trained using \(\mathcal{L_{\text{objects}}}\), while the baseline model is frozen.

\noindent{\paragrax{Baseline model supervision.}} The baseline model is supervised based on the predictions of the decoder with the ground truth grid \(\mathbf{Y}_t \in \{c_0, c_1, \dots, c_M\}^{L \times W \times H}\) through the Semantic Scene Completion (\textsc{ssc}) loss \(\mathcal{L_{\textsc{ssc}}}\) \cite{Cao2022}.

\noindent{\paragrax{Object module supervision.}} 
The computation of the objects loss \(\mathcal{L_{\text{objects}}}\) stems from two components: an object detection loss \(\mathcal{L_{\text{det}}}\) and an occupancy loss \(\mathcal{L}_{\text{occ}}\).

The object detection loss is similar to the approach in \textsc{detr} \cite{carion2020end}, which combines a classification loss \(\mathcal{L}_{\text{cls}}\) and a center distance loss \(\mathcal{L}_{\text{dist}}\). We use a focal loss \cite{8237586} for the classification and \(L_2\) loss for the distance loss, resulting in

\begin{equation}\label{eq:obj_det_loss}
    \mathcal{L_{\text{det}}} = \lambda_{\text{1}}\mathcal{L}_{\text{cls}} + \lambda_{\text{2}}\mathcal{L}_{\text{dist}},
\end{equation}

We match object predictions to ground truth with the Hungarian algorithm \cite{kuhn1955hungarian}, using a cost function analogous to the loss in \cref{eq:obj_det_loss} (any unmatched predictions are associated with the background class), which defines a mapping \(\sigma_{\text{det}}\) that assigns object prediction indices \(i\) to ground truth indices \(j\) such that \(j = \sigma_{\text{det}}(i)\) (where the ground truth has been padded with no-object entries to allow queries to predict no object).

For the occupancy loss, once the association at the object level is established, we apply the same concept again at the voxel level, within each individual object. Let us consider a prediction–ground truth pair \(((\hat{y}^{i}, \mathbf{\hat{c}}^{i}, \hat{V}^{i}); (y^{\sigma_{\text{det}}(i)}, \mathbf{c}^{\sigma_{\text{det}}(i)}, V^{\sigma_{\text{det}}(i)}))\). We want to find the best match between the predicted point cloud \({\hat{V}}^{i}\) and the list of ground truth voxel centers \(V^{\sigma_{\text{det}}(i)}\) occupied by the object. This can be achieved by reapplying the Hungarian algorithm (bipartite matching) at the voxel level. Assuming \(K\) is always larger than the number of voxels occupied by a single object, we convert the ground truth set to a list of \(K\) elements by padding with \(\varnothing\).
The matching cost function for a single offset–ground truth voxel pair is given by

\begin{equation}
    \begin{aligned}
        \mathcal{L}_{\text{match}}(k, \sigma(k)) = & - \indic{s_{\sigma(k)} \neq \varnothing} \hat{s}_k\\
        & + \indic{s_{\sigma(k)} \neq \varnothing} \mathcal{L}_{\text{dist}} \left( \mathbf{\hat{v}}_k, \mathbf{v}_{\sigma(k)} \right)
    \end{aligned}
\end{equation}

\noindent where \(\sigma\) is one of the possible matching permutations. Here, we also use the \(L_2\) norm as a distance cost between the predicted offset and the matched voxel center.

Once the optimal match \(\sigma_{\text{occ}}\) is determined, we compute the loss as

\begin{equation}
    \mathcal{L}_{\text{occ}} = \sum_{k=1}^K\left[-\log\hat{s}_k+\indic{s_{\sigma_{\text{occ}}(k)} \neq \varnothing} \mathcal{L}_{\text{dist}}(\mathbf{\hat{v}}_k, \mathbf{v}_{\sigma_{\text{occ}}(k)})\right].
\end{equation}
In practice, we also use a focal loss for the occupancy \(\hat{s}\) and balance the classification and regression components with coefficients, which leads to

\begin{equation}
    \mathcal{L_{\text{occ}}} = \lambda_{\text{3}}\mathcal{L_{\text{focal}}} + \lambda_{\text{4}}\mathcal{L}_{\text{dist}}.
\end{equation}

Finally, the object loss is given by

\begin{equation}
    \mathcal{L_{\text{objects}}} = \mathcal{L_{\text{det}}} + \mathcal{L_{\text{occ}}}.
\end{equation}

\noindent{\paragrax{Location and point cloud supervision decoupling.}} During training, unlike during inference, we add the predicted offsets to the \emph{ground truth location} \(\mathbf{c}^{i}\) of the matched object, instead of the predicted location \(\mathbf{\hat{c}}^{i}\). This approach decouples the learning of object centers and offsets, ensuring that the offset loss does not depend on the accuracy of the predicted objects' locations. As a result, the model focuses on learning the actual object occupancy without compensating for inaccurate location predictions, allowing each component to learn independently.

Lastly, we only train on objects that have at least one visible voxel from the cameras, even though they may occupy several non-visible voxels.

%% file: images/architecture.tex
\begin{figure*}[htbp]
    \includegraphics[width=\textwidth]{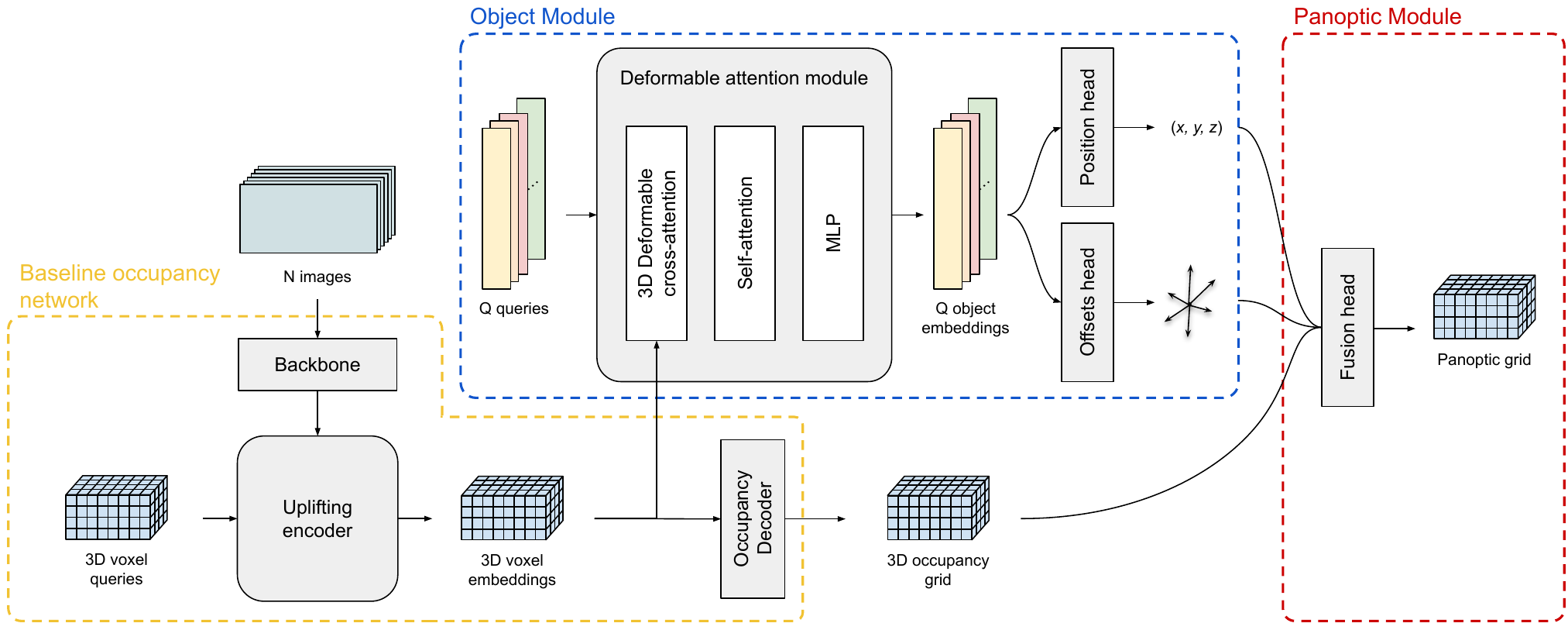}
    \caption{\textbf{Overview of our framework \ourmodelname{}.} Our approach can be integrated with modern occupancy models and consists of two key components: an \textit{Object Module} and a \textit{Panoptic Module}. The \emph{Object Module} utilizes features extracted by an uplifting encoder to detect objects in the voxel grid and predict their occupancy using the offset occupancy mechanism. The \emph{Panoptic Module} then merges these object predictions with the occupancy grid output to produce a panoptic occupancy map. Classification head in the \emph{Object Module} omitted for conciseness.}
    \label{fig:architecture}
\end{figure*}

%% file: sec/4_experiment.tex
\section{Experiments}
\label{sec:experiments}

\subsection{Datasets}

\noindent{\textbf{nuScenes dataset}}. nuScenes \cite{nuscenes2019} is a large-scale, multimodal dataset containing 1,000 driving scenes, divided into 700 scenes for training, 150 for validation, and 150 for testing. Each scene is approximately 20 seconds long and recorded at a frequency of 20 Hz. The dataset provides \textsc{rgb} images captured from six cameras with a \(360^{\circ}\) horizontal field of view, along with \textsc{lidar} point cloud data. For the object detection task, nuScenes includes annotations of 3\textsc{d} bounding boxes on keyframes at 2 Hz, labeled with 10 \textit{thing} classes.

\noindent{\textbf{Occ3D-nuScenes}}. Occ3D-nuScenes \cite{Tian2023} extends the nuScenes dataset by incorporating voxel-wise semantic occupancy annotations, making it suitable for 3\textsc{d} occupancy prediction. The occupancy grid covers a range of \(-40\,\text{m}\) to \(40\,\text{m}\) along the \(x\) and \(y\) axes, and from \(-1\,\text{m}\) to \(5.4\,\text{m}\) along the \(z\)-axis in the ego coordinate frame. The voxel grid is defined with a resolution of \(0.4\,\text{m} \times 0.4\,\text{m} \times 0.4\,\text{m}\), allowing for dense spatial representation. Occ3D-nuScenes provides semantic labels for 17 categories, which include 16 known object classes and an additional \say{empty} class. Additionally, it includes a visibility mask to indicate occluded voxels in the camera view. Since the Occ3D-nuScenes dataset does not include panoptic annotations, we generate them offline by intersecting semantic voxel annotations with 3\textsc{d} bounding boxes.

\noindent{\paragrax{Evaluation metrics.}} We use \ac{iou} to evaluate scene occupancy by distinguishing between occupied and non-occupied voxels, independent of semantic labels. This binary occupancy map is essential for capturing spatial structure in 3\textsc{d} environments. To assess semantic segmentation performance, we compute \ac{miou} over semantic classes, accounting for both occupancy and class-specific accuracy. The standard evaluation protocol for 3\textsc{d} semantic occupancy relies on the visibility mask, computing errors only on visible voxels. However, this approach overlooks the accuracy of object shape reconstruction, which is crucial for safe path planning and aligns more naturally with a panoptic framework. To provide a more comprehensive evaluation and facilitate future comparisons, we also report performance without applying the visibility mask. We also report the Panoptic Quality (\textsc{pq}) metric for the panoptic \textsc{lidar} segmentation task. \textsc{pq} is defined as the product of Recognition Quality (\textsc{rq}), which measures detection performance, and Segmentation Quality (\textsc{sq}), which evaluates the segmentation consistency of correctly matched instances \cite{kirillov2019panoptic}.

\subsection{Experimental settings.}\label{sec:experiments-implementation}

The input images are \(1600 \times 1200\) and augmented with GridMask \cite{chen2020gridmask} and photometric distortion.

For the baseline model, we take inspiration from \cite{Tong2023, Li2022}. We use a ResNet101 \cite{he2016deepResNet} pre-trained on ImageNet \cite{conf-cvpr-DengDSLL009} and an \textsc{fpn} \cite{lin2017feature} as the backbone. We implement the encoder with a transformer that uses a grid of 1\textsc{d} sinusoidal positional voxel queries \(\mathbf{Q} \in [-1, 1]^{L/2 \times W/2 \times H/2 \times D}\), and a stack of four layers with deformable cross-attention \cite{zhu2020deformable} and 3\textsc{d} deformable self-attention \cite{Tong2023, zhu2020deformable}. The cross-attention layers project reference points (the voxel grid centers) into camera views via projection matrices, aggregating image features potentially from multiple views (in which case features for that point are averaged). The hidden network dimension \(D\) is 128, and the voxel embeddings are predicted in a grid at half resolution. The decoder upsamples the voxel embeddings to the output resolution of \(200 \times 200 \times 16\) using trilinear interpolation and predicts the class per voxel with a linear layer.

In the \emph{Object Module}, we set the number of object queries \(Q = 900\), the number of layers \(M = 4\), and the number of offsets \(K = 2197\). In the \emph{Panoptic Module}, we set \(r = 9\) and use the Manhattan distance.

The loss weights are set as follows: \(\lambda_1 = 0.5\), \(\lambda_2 = 0.02\), \(\lambda_3 = 0.125\), \(\lambda_4 = 0.0125\). Each stage runs for 50 epochs with AdamW \cite{loshchilov2018decoupled}, a learning rate of \(2 \times 10^{-4}\), and an exponential learning rate decay scheduler.

% \vspace{-3pt}
\subsection{Main results}
% \vspace{-2pt}
In this section we evaluate the performance of our model for the task of 3\textsc{d} semantic occupancy and panoptic \textsc{lidar} segmentation as a proxy for panoptic occupancy performance. In \cref{fig:qualitative_examples} we include qualitative results of our model.

\noindent{\paragrax{3\textsc{d} semantic occupancy.}} We evaluate our model on the nuScenes validation set, with results summarized in \cref{tab:camera_occ_short}. The model achieves 28.0 \ac{miou} and 43.9 \ac{iou}, delivering performance comparable to existing works. Since our goal is to develop a model capable of predicting complete object occupancies, we also report, in \cref{tab:camera_occ_short}, the model's performance evaluated without the visibility mask, providing a more comprehensive assessment of its reconstruction capabilities. Under these settings, it achieves 17.2 \ac{miou} and 24.9 \ac{iou}. For completeness, we also report the performance of previous work trained with the visibility mask.

\noindent{\paragrax{Panoptic \textsc{lidar} segmentation.}} Results on panoptic \textsc{lidar} segmentation are reported in \cref{tab:lidar_panoptic}. For evaluation, \textsc{lidar} points were labeled according to their corresponding voxel labels in the occupancy grid. The model achieved a \textsc{pq} score of 29.4, demonstrating the validity of our approach. It is likely that the gap with the best-performing models can be attributed to two main factors. First, our framework does not retrain the base model, which might limit the relevant features that it can extract; second, our method does not incorporate temporal consistency across frames, which could be especially useful for improving predictions of distant and (partially) occluded objects.

\input{images/qualitative_image}

\input{./table/nuscenes_occ_val_short}
\input{./table/nuscenes_val_panoptic}

\subsection{Discussion and ablation}
\label{sec:discussion}

In this section, we analyze the impact of our offsets loss and location error decoupling, the computational cost introduced by the \emph{Object Module} for panoptic occupancy, and evaluate the effect of the majority voting radius \(r\) in the \emph{Panoptic Module}.

\input{./table/design_choices}

\noindent{\paragrax{Offsets loss and location error decoupling.}} Since the offsets \(L_2\) loss focuses on reconstructing the object point cloud as accurately as possible, it is more effective, during supervision, to compare the predicted point cloud with the ground truth point cloud without factoring in the localization error of the object center. As discussed in \cref{sec:train_methodology}, during supervision we sum the predicted offsets to the ground truth object center. The loss between the resulting point cloud and the ground truth point cloud will therefore only penalize inaccurate offsets. We compare, in \cref{tab:ablation}, the proposed model with a model that does not employ this design choice; \ie during supervision, the point cloud is determined as the sum of the predicted object location and predicted offsets. We compare two models that use voting radius \(r = 0\) to evaluate the raw offsets' quality. The results indicate that decoupling produces better object segmentation, achieving 17.0 \textsc{pq}$^{Th}$ compared to 15.8 \textsc{pq}$^{Th}$ for the model without the decoupling strategy. Notably, decoupling improves \textsc{rq}$^{Th}$ by enhancing the shape of predicted objects, resulting in a higher number of true positives, which validates the effectiveness of our design choice.

\noindent{\paragrax{Panoptic module voting radius.}} We analyze the impact of the majority voting radius \(r\), as described in \cref{sec:architecture}, using panoptic metrics. As shown in \cref{fig:panoptic_head_threshold}, performance initially improves as \(r\) increases, as a larger radius helps to smooth out predictions. However, beyond the optimal value of 9, performance declines because a larger \(r\) leads to incorrect \textsc{id} assignments, as the majority voting starts considering irrelevant voxels. Notably, \textsc{pq}$^{Th}$ increases mostly due to \textsc{rq}$^{Th}$, as the majority voting mechanism smooths predictions and therefore increases the number of true positives.

\input{images/manhattan_distance}

\noindent{\paragrax{\ourmodelname{} complexity.}} \Cref{tab:object_head_overhead} presents an analysis of the overhead introduced by our framework. Our framework introduces minimal overhead, making it an efficient extension to existing 3\textsc{d} semantic occupancy models. The parameter increase is negligible, with only 2.3 million (\textsc{m}) additional parameters. During training, memory consumption remains low, as we freeze the base model's weights and optimize only the \emph{Object Module}, significantly reducing computational demands. During evaluation, memory usage increases due to the voting mechanism in the \emph{Panoptic Module}. Despite this, the overall latency increase is modest, allowing the model to maintain efficiency for real-world deployment.

\input{./table/panoptic_head_complexity}

%% file: images/qualitative_image.tex
\input{misc/occupancy_colors}
\begin{figure*}[htbp]

    \begin{minipage}[b]{\textwidth}
    \centering
        \scriptsize
        \begin{tabular}{c@{\hskip 5.0pt}c@{\hskip 5.0pt}c@{\hskip 5.0pt}c@{\hskip 5.0pt}c@{\hskip 5.0pt}c@{\hskip 5.0pt}c@{\hskip 5.0pt}c@{\hskip 5.0pt}c@{\hskip 5.0pt}c@{\hskip 5.0pt}c@{\hskip 5.0pt}c@{\hskip 5.0pt}c@{\hskip 5.0pt}c@{\hskip 5.0pt}c@{\hskip 5.0pt}c@{\hskip 5.0pt}}
            \textcolor{nbarrier}{$\blacksquare$} & \textcolor{nbicycle}{$\blacksquare$} & \textcolor{nbus}{$\blacksquare$} & \textcolor{ncar}{$\blacksquare$} & \textcolor{nconstruct}{$\blacksquare$} & \textcolor{nmotor}{$\blacksquare$} & \textcolor{npedestrian}{$\blacksquare$} & \textcolor{ntraffic}{$\blacksquare$} & \textcolor{ntrailer}{$\blacksquare$} & \textcolor{ntruck}{$\blacksquare$} & \textcolor{ndriveable}{$\blacksquare$} & \textcolor{nother}{$\blacksquare$} & \textcolor{nsidewalk}{$\blacksquare$} & \textcolor{nterrain}{$\blacksquare$} & \textcolor{nmanmade}{$\blacksquare$} & 
            \textcolor{nvegetation}{$\blacksquare$} \\
            barrier & bicycle & bus & car & const. veh. & motorcycle & pedestrian & traffic cone & trailer & truck & drive. surf. & other flat & sidewalk & terrain & manmade & vegetation
        \end{tabular}
    \vspace{2pt}

    \captionsetup[subfigure]{labelformat=empty}
    \centering
    \begin{subfigure}[b]{0.45\textwidth}
        \centering
        \includegraphics[width=\textwidth]{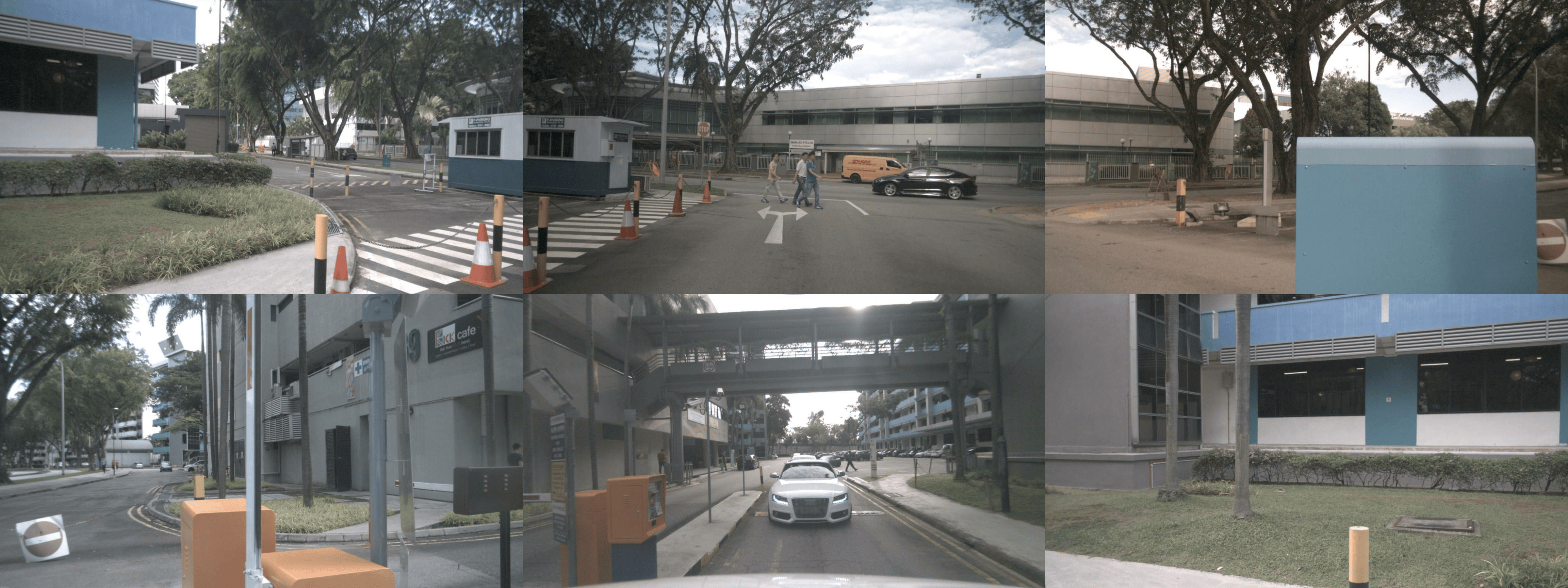}
    \end{subfigure}
    \hspace{1pt}
    \begin{subfigure}[b]{0.17\textwidth}
        \centering
        \includegraphics[width=\textwidth]{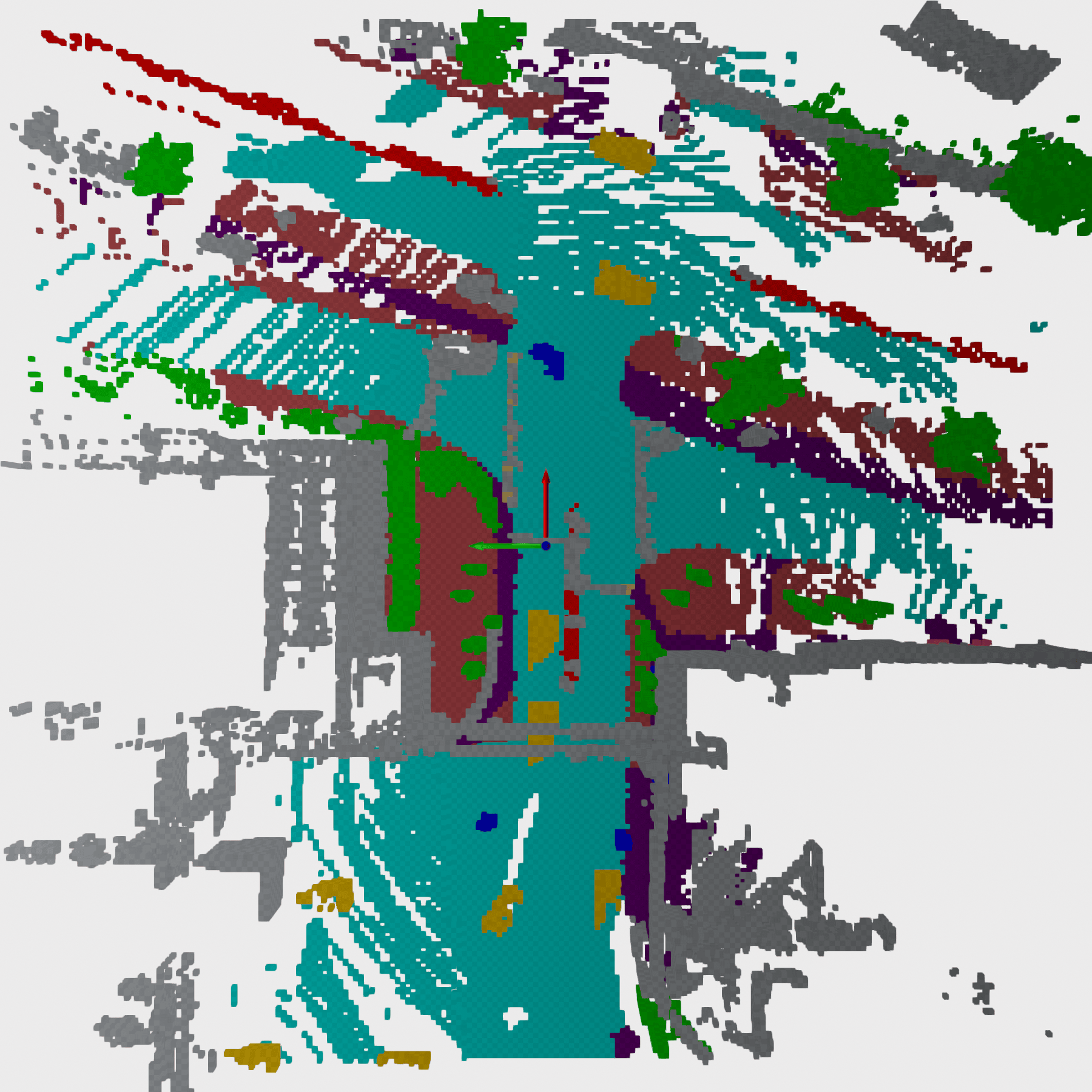}
    \end{subfigure}
    \hspace{1pt}
    \begin{subfigure}[b]{0.17\textwidth}
        \centering
        \includegraphics[width=\textwidth]{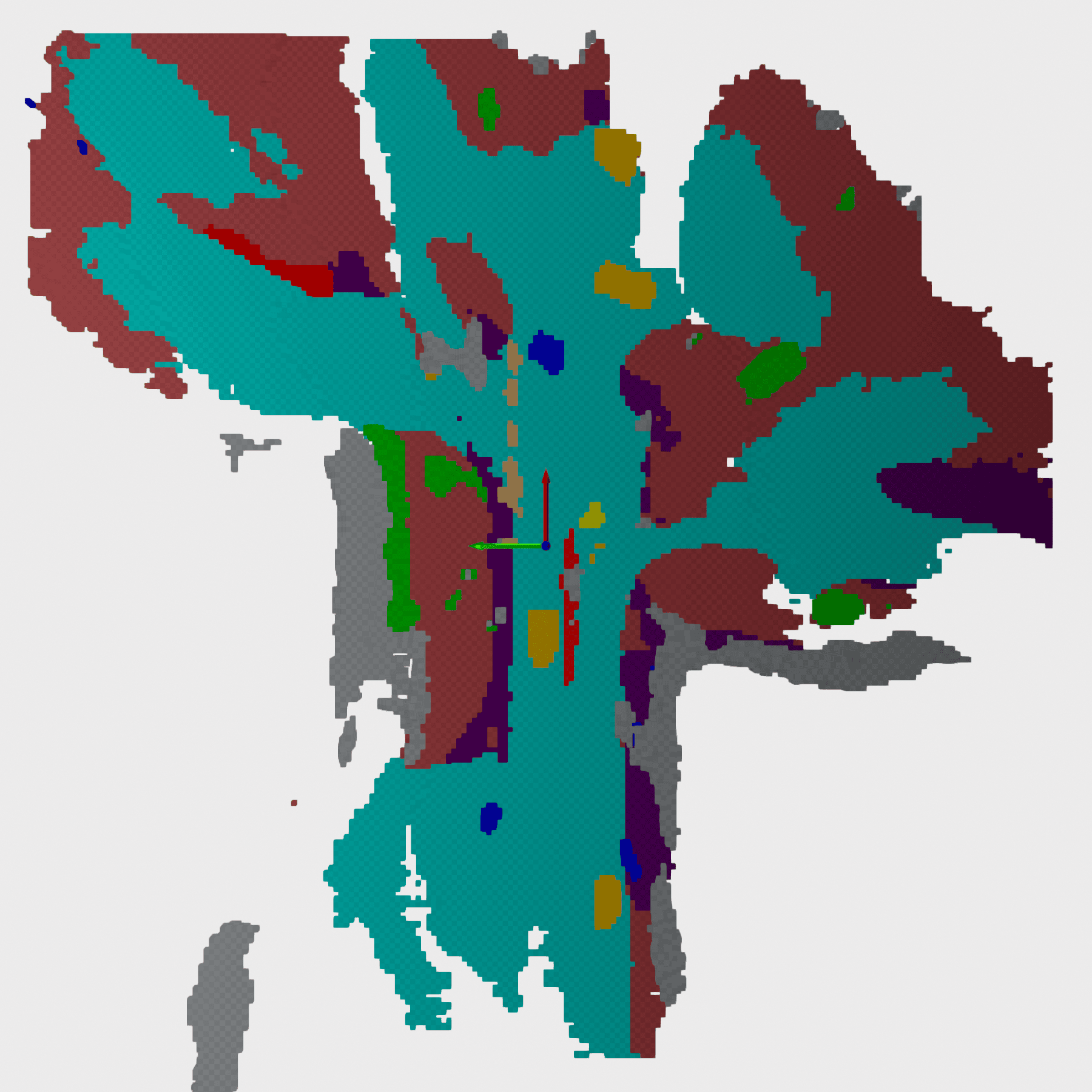}
    \end{subfigure}
    \hspace{1pt}
    \begin{subfigure}[b]{0.17\textwidth}
        \centering
        \includegraphics[width=\textwidth]{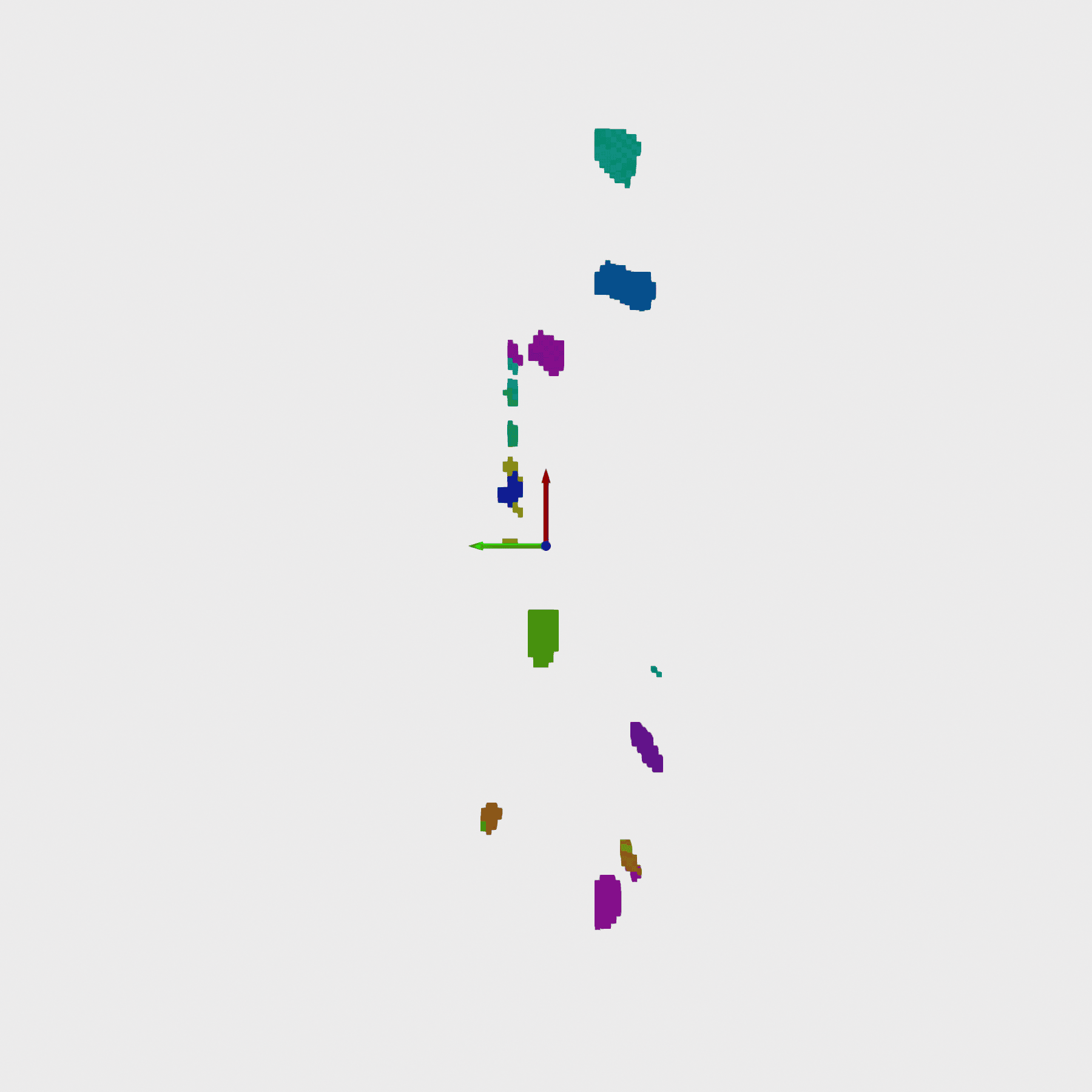}
    \end{subfigure}

    \vspace{3pt}
    
    \begin{subfigure}[b]{0.45\textwidth}
        \centering
        \includegraphics[width=\textwidth]{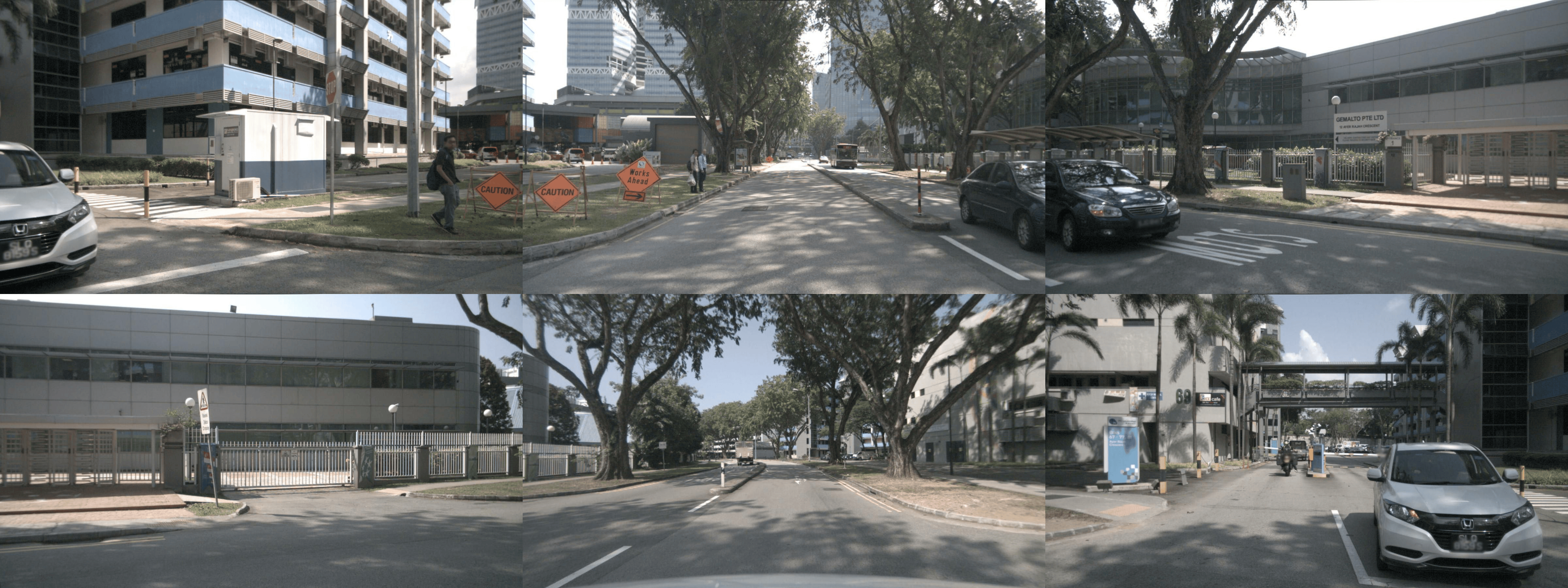}
        \caption{\textbf{Multi-View Images}}
    \end{subfigure}
    \hspace{1pt}
    \begin{subfigure}[b]{0.17\textwidth}
        \centering
        \includegraphics[width=\textwidth]{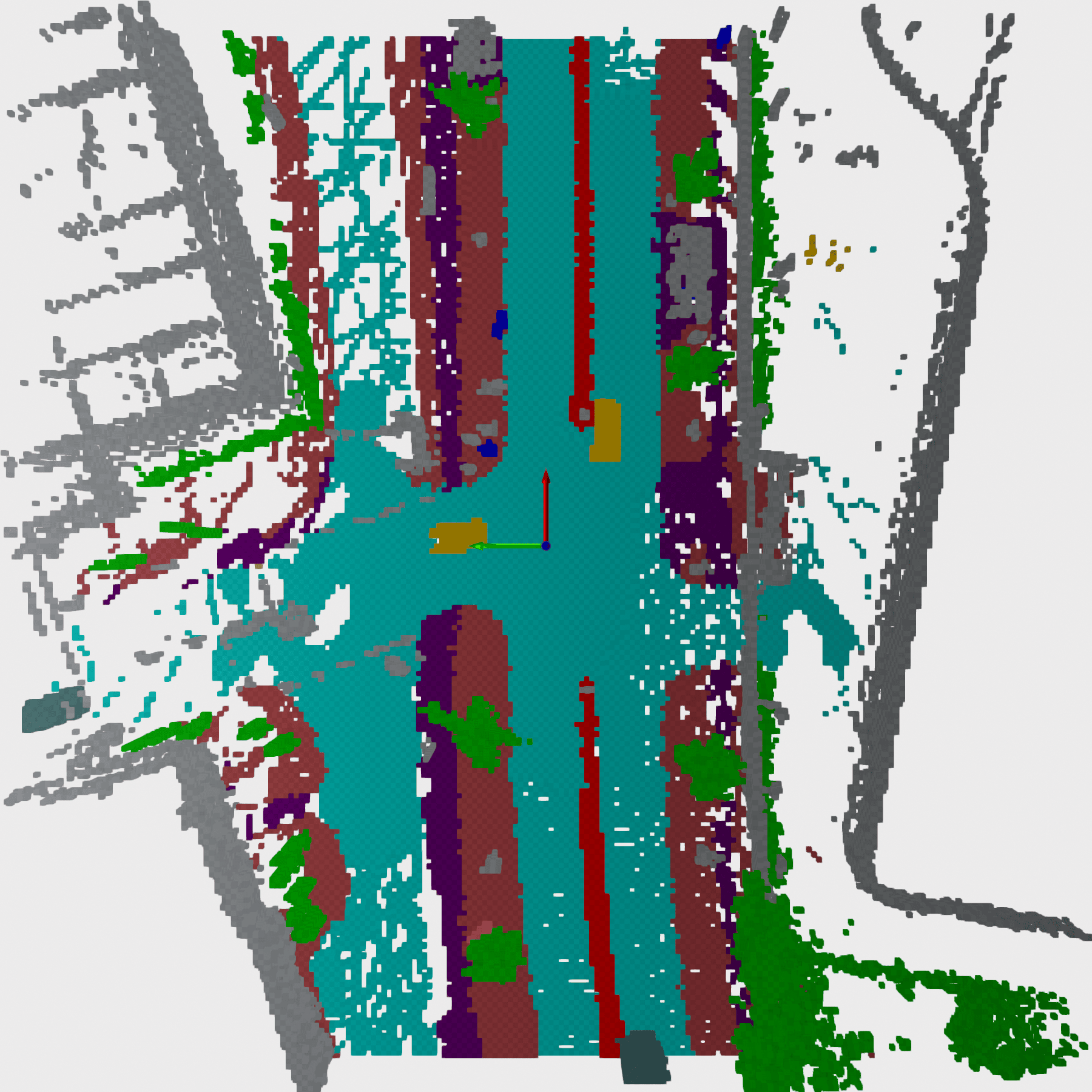}
        \caption{\textbf{GT occupancy}}
    \end{subfigure}
    \hspace{1pt}
    \begin{subfigure}[b]{0.17\textwidth}
        \centering
        \includegraphics[width=\textwidth]{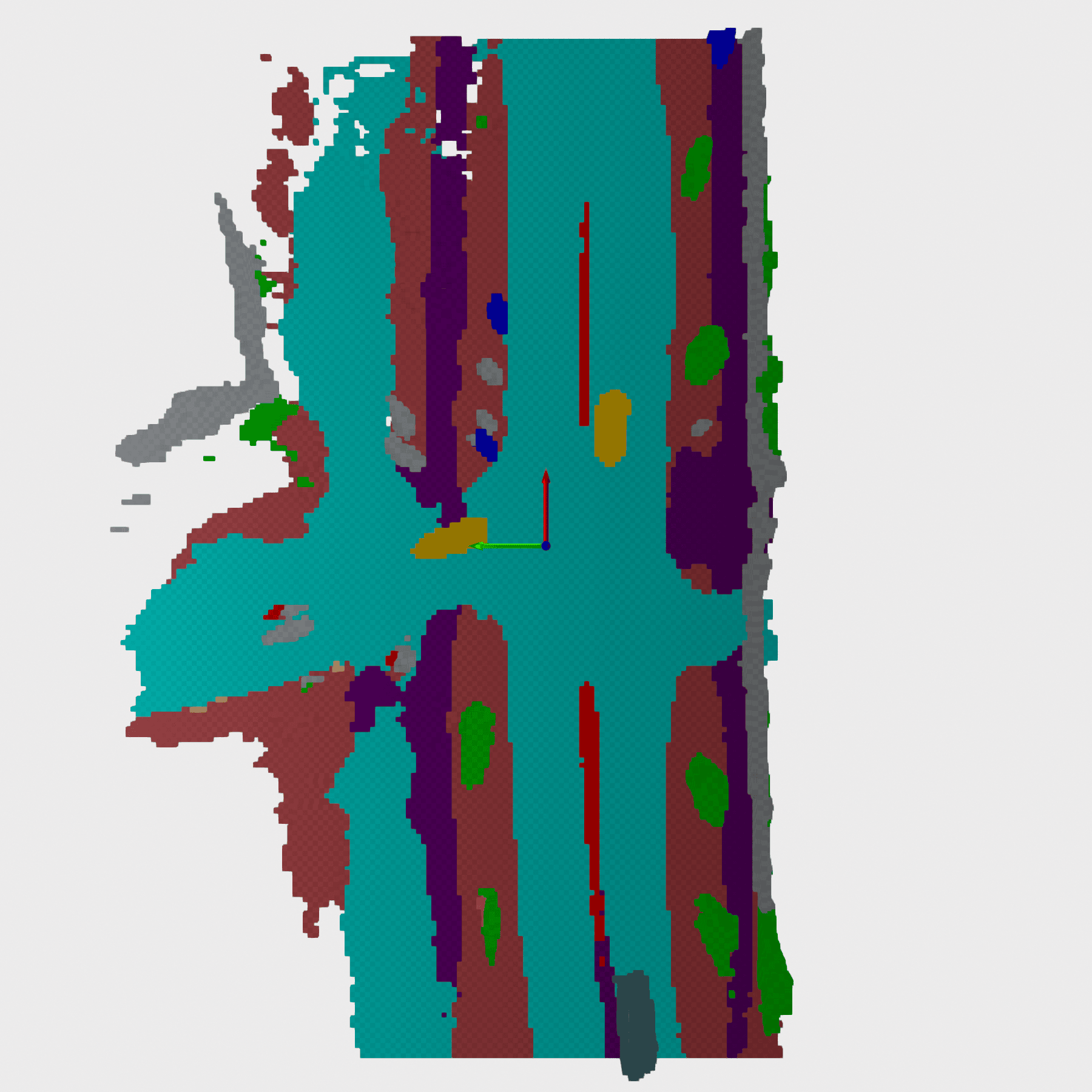}
        \caption{\textbf{Pred occupancy}}
    \end{subfigure}
    \hspace{1pt}
    \begin{subfigure}[b]{0.17\textwidth}
        \centering
        \includegraphics[width=\textwidth]{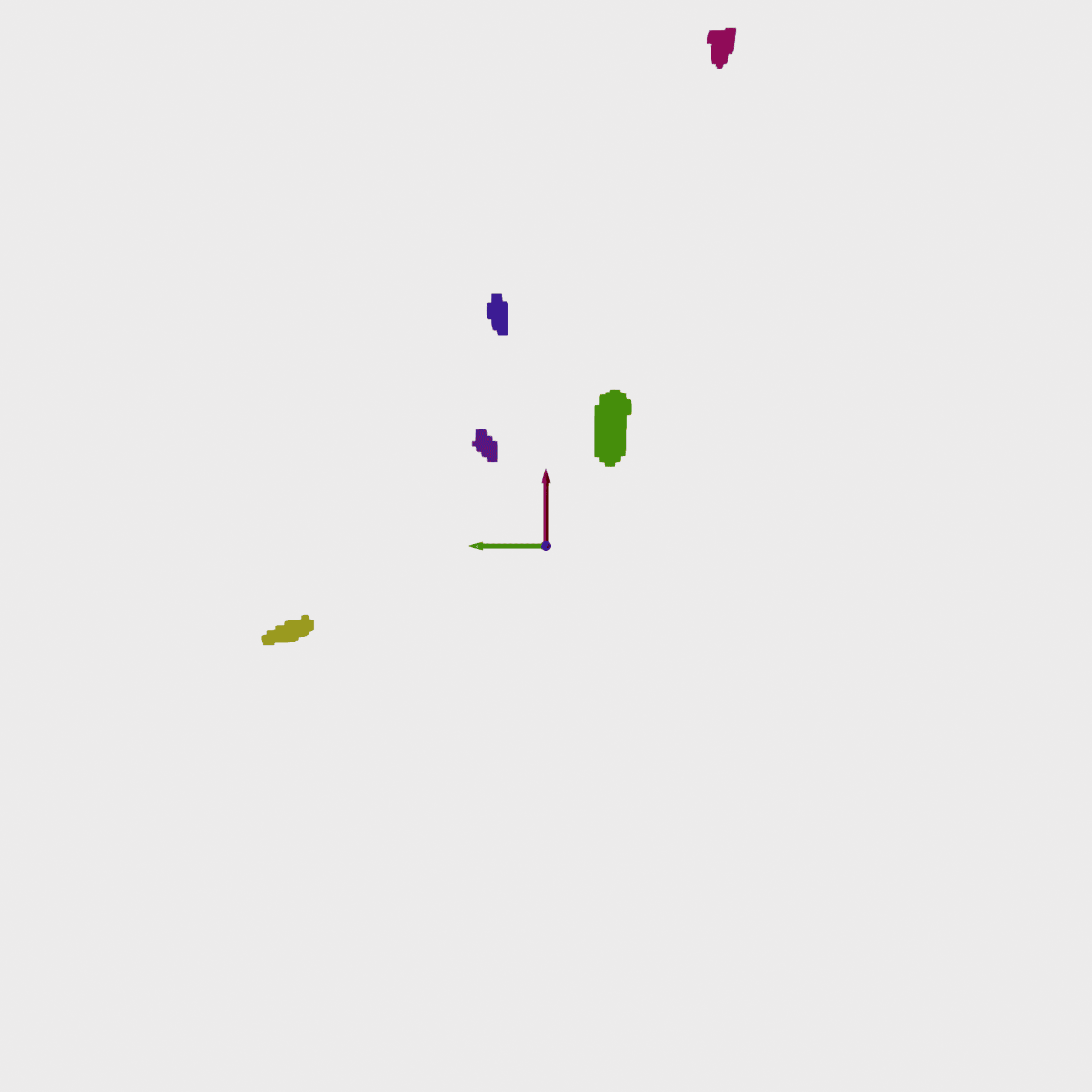}
        \caption{\textbf{Pred panoptic}}
    \end{subfigure}

    \end{minipage}
    
    \caption{\textbf{Qualitative results on Occ3D-nuScenes validation set.} Our model accurately differentiates and segments different objects in the scene.}
    
    \label{fig:qualitative_examples}
\end{figure*}

%% file: table/nuscenes_occ_val_short.tex
\input{misc/occupancy_colors}
\begin{table}[ht]
	\footnotesize
 	\setlength{\tabcolsep}{0.005\linewidth}
	
    \begin{center}

	\begin{tabular}{l | c c c c | r r}
		\toprule
		Method & \makecell{Image \\ Backbone} & Temporal & \makecell{Train \\ w/ mask} & \makecell{Evaluate \\ w/ mask} & mIoU & IoU \\
		\midrule
        MonoScene~\cite{Cao2022}                & R101-DCN  & \xmark & \xmark & \cmark &  6.1 & - \\
        BEVDet~\cite{Huang2021}                 & R101-DCN  & \xmark & \xmark & \cmark & 19.4 & - \\
        OccFormer \cite{Zhang2023}              & R101      & \xmark & \xmark & \cmark & 21.9 & - \\
        BEVFormer~\cite{Li2022}                 & R101-DCN  & \cmark & \xmark & \cmark & 26.9 & - \\
        TPVFormer~\cite{Huang2023}              & R101-DCN  & \cmark & \xmark & \cmark & 27.8 & - \\
        CTF-Occ~\cite{Tian2023}                 & R101-DCN  & \xmark & \xmark & \cmark & 28.5 & - \\
        SparseOcc~\cite{Liu2023}                & R50       & \cmark & \xmark & \cmark & 30.9 & - \\
        PanoOcc~\cite{Wang2024PanoOcc}          & R101-DCN  & \cmark & \xmark & \cmark & 32.5 & - \\
        \midrule
        TPVFormer$\ddagger$~\cite{Huang2023}    & R50       & \cmark & \cmark & \cmark & 34.2              & 66.8 \\
        OccFormer$\ddagger$~\cite{Zhang2023}    & R50       & \xmark & \cmark & \cmark & 37.4              & \underline{70.1} \\ 
        BEVFormer~\cite{Li2022}                 & R101-DCN  & \cmark & \cmark & \cmark & \underline{39.2}  & -     \\
        PanoOcc~\cite{Wang2024PanoOcc}          & R101-DCN  & \cmark & \cmark & \cmark & \textbf{44.5}     & \textbf{75.0} \\
        \midrule
        \textbf{\ourmodelname{} (Ours)}         & R101      & \xmark & \xmark & \cmark & 28.0 & 43.9 \\
        \textbf{\ourmodelname{} (Ours)}         & R101      & \xmark & \xmark & \xmark & 17.2 & 24.9 \\
	\bottomrule
	\end{tabular}
    \end{center}
    \caption{\textbf{3D Occupancy prediction performance on the Occ3D-nuScenes dataset.} \say{Temporal} indicates that the model uses past frames when generating predictions. \say{Train w/ mask} and \say{Evaluate w/ mask} indicate whether the model has been trained using the camera mask and whether the performance has been measured using the camera mask, respectively. $\ddagger$ indicates performance measured by \cite{Ma2024}. Best performance is \textbf{bolded} and second best is \underline{underlined}. Per-class performance can be found in the supplementary tables in \cref{sec:results-supp}.}
    \label{tab:camera_occ_short}
    \vspace{-10pt}
\end{table}

%% file: table/nuscenes_val_panoptic.tex
\begin{table}[t]
    \small

    \def\b{\textbf}
    \def\u{\underline}
    
    \begin{center}
     \setlength{\tabcolsep}{0.01\linewidth}
    \begin{tabular}{c|c|cccc}
		\toprule
		Method &  \makecell{Input \\ Modality} & PQ & PQ$^{\dagger}$ & RQ & SQ \\
	\midrule
        EfficientLPS~\cite{sirohi2021efficientlps}  & \textsc{lidar}  &    62.0  &    65.6  &    73.9  &    83.4  \\
        Panoptic-PolarNet~\cite{zhou2021panoptic}   & \textsc{lidar}  &    63.4  & \u{67.2} &    75.3  &    83.9  \\
        Panoptic-PHNet~\cite{li2022panoptic}        & \textsc{lidar}  & \u{74.7} & \b{77.7} & \u{84.2} & \u{88.2} \\
        LidarMulitiNet~\cite{ye2022lidarmultinet}   & \textsc{lidar}  & \b{81.8} &       -  & \b{90.8} & \b{89.7} \\
        \midrule
        PanoOcc~\cite{Wang2024PanoOcc}              & Camera &    62.1  &    66.2  &    75.1  &    82.1  \\
        \textbf{\ourmodelname{} (Ours)}             & Camera &    29.4  &    38.3  &    40.1  &    66.5  \\
	\bottomrule
	\end{tabular}
    \end{center}
    \vspace{-12pt}
    \caption{\textbf{\textsc{lidar} panoptic segmentation results on nuScenes validation set.} Our \ourmodelname{} exhibits reasonable performance, albeit clearly behind the state of the art (\textsc{sota}). Many recent improvements included in the \textsc{sota} can readily be included in our method which will likely reduce the gap significantly.}
    \label{tab:lidar_panoptic}
    \end{table}

%% file: table/design_choices.tex
\begin{table}[]
\vspace{15pt}
\begin{center}
\begin{tabular}{l|cccc}
\toprule
Variation  & PQ$^{Th}$ & RQ$^{Th}$ & SQ$^{Th}$ \\
\midrule
\ourmodelname{} w/o decoupling      &         15.8  &         22.0  & \textbf{69.9} \\
\ourmodelname{}                     & \textbf{17.0} & \textbf{23.7} &         69.7 \\     
\bottomrule
\end{tabular}
\end{center}
\vspace{-12pt}
\caption{\textbf{Offsets loss and location error decoupling.} Decoupling object location error from point cloud supervision improves performance on object classes, validating the effectiveness of our approach. The models were trained on 50\% of the dataset and evaluated on the full validation set of nuScenes.}
\label{tab:ablation}
\end{table}

%% file: images/manhattan_distance.tex
\begin{figure}[h]
    \includegraphics[width=\columnwidth]{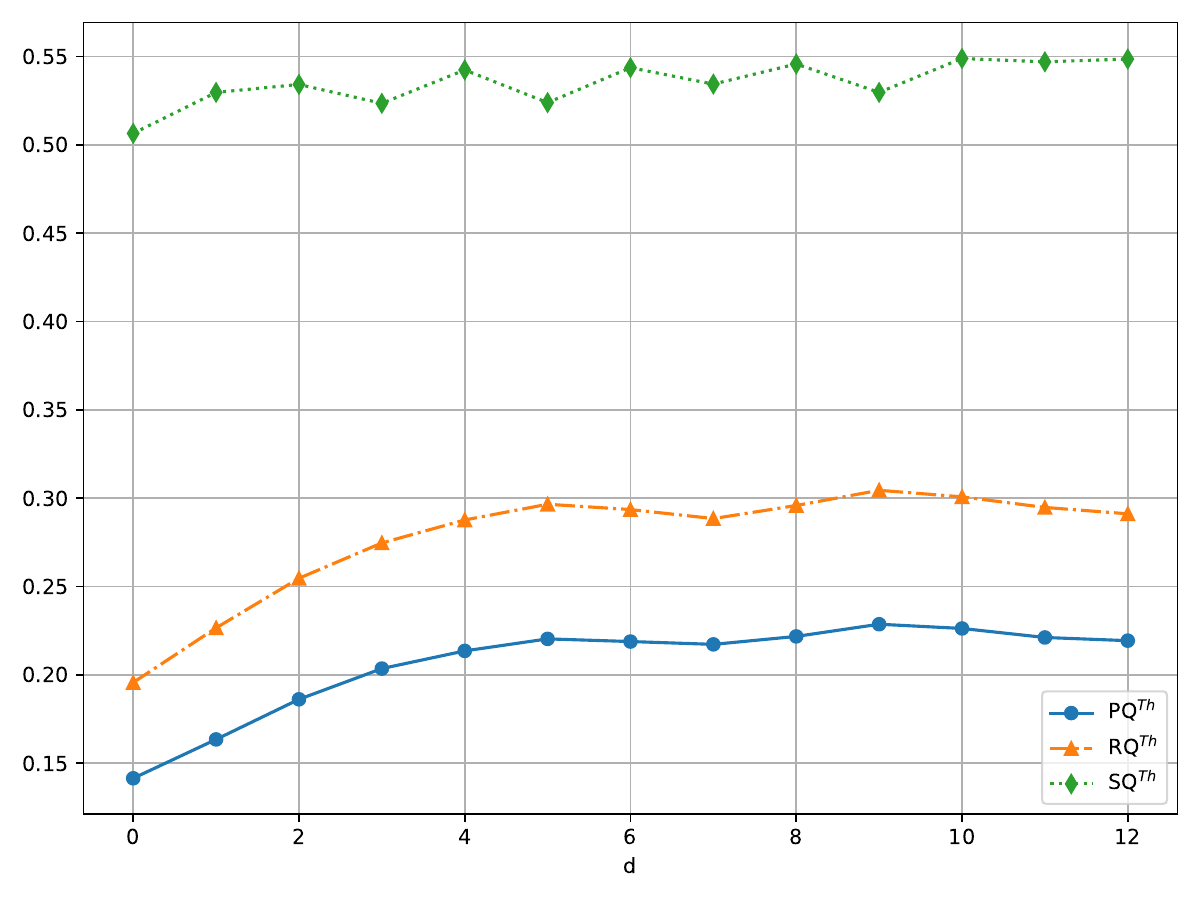}
    \vspace{-20pt}
    \caption{\textbf{Effect of the voting radius.} Increasing $r$ initially improves performance by smoothing predictions. However, beyond the optimal value of $9$, performance declines as the voting process starts to incorporate irrelevant voxels, leading to incorrect \textsc{id} assignments. Tested on the mini set of nuScenes.}
    \label{fig:panoptic_head_threshold}
\end{figure}

%% file: table/panoptic_head_complexity.tex
\begin{table}[hbt!]
    \small
    \begin{center}
     \setlength{\tabcolsep}{0.015\linewidth}
    \begin{tabular}{l|cccc}
		\toprule
		  Method & Params & Memory & Latency  & FPS \\
	    \midrule
            Baseline & 44.2M & 14.4G/2.5G & 500 ms  & 2.0 \\
            Baseline w/ \ourmodelname{} & 46.5M & 2.2G/14.0G   & 714 ms & 1.4 \\
		\bottomrule
    \end{tabular}
 \end{center}
 \vspace{-12pt}
\caption{\textbf{Overhead of our framework \ourmodelname{}.} Our framework is lightweight in both parameter count and latency while enabling panoptic occupancy. We show train/inference memory. Each frame includes 6 cameras. Tested on the Nvidia L40S \textsc{gpu}.}
    \label{tab:object_head_overhead}
    \vspace{-10pt}
\end{table}

%% file: sec/5_conclusion.tex
\section{Conclusion}
\label{sec:conclusion}

In this work, we propose a novel approach to 3\textsc{d} panoptic scene completion. In particular, we introduce a framework that leverages annotated occupancy maps to learn object shapes, a crucial aspect for path planning in autonomous driving contexts. Our method identifies objects and estimates their occupancy using a set of offsets relative to their positions. This approach enables the formulation of object shape learning as a continuous and differentiable problem.

\noindent{\paragrax{Limitations.}} The sequential nature of the Hungarian algorithm applied at the object level introduces a bottleneck and increases the total training time. This issue is partially mitigated by increasing the number of training processes, such as using multiple \textsc{gpu}s, where each process handles a smaller subset of objects.

\noindent{\paragrax{Future work.}} Our method highlights the potential of modeling object shapes as a continuous and differentiable problem, laying a foundation for future work in this area. Future work will focus on integrating temporal reasoning to improve frame-to-frame consistency, as well as refining hyperparameters and designing a more comprehensive \emph{Panoptic Module} to enhance panoptic performance.

\vspace{10pt}

\paragrax{Acknowledgments.} We would like to thank Pierre-François De Plaen for his meaningful insights and comments. The resources and services used in this work were provided by the VSC (Flemish Supercomputer Center), funded by the Research Foundation - Flanders (FWO) and the Flemish Government. We greatly acknowledge the Flanders AI Research program.

%% file: sec/X_suppl.tex
\maketitlesupplementary

\section{Detailed results and additional qualitative examples}
\label{sec:results-supp}

This section provides more detailed results on the Occupancy task of Occ3D-nuScenes. \Cref{tab:camera_occ_things,tab:camera_occ_stuff} provide IoU numbers for each individual \emph{things} and \emph{stuff} class, respectively. We also provide additional qualitative results in \Cref{fig:qualitative_examples_supp}.

\include{images/qualitative_image_supp}

\include{table/nuscenes_occ_val}

%% file: images/qualitative_image_supp.tex
\input{misc/occupancy_colors}
\begin{figure*}[]

    \begin{minipage}[b]{\textwidth}
    \centering
        \scriptsize
        \begin{tabular}{c@{\hskip 5.0pt}c@{\hskip 5.0pt}c@{\hskip 5.0pt}c@{\hskip 5.0pt}c@{\hskip 5.0pt}c@{\hskip 5.0pt}c@{\hskip 5.0pt}c@{\hskip 5.0pt}c@{\hskip 5.0pt}c@{\hskip 5.0pt}c@{\hskip 5.0pt}c@{\hskip 5.0pt}c@{\hskip 5.0pt}c@{\hskip 5.0pt}c@{\hskip 5.0pt}c@{\hskip 5.0pt}}
            \textcolor{nbarrier}{$\blacksquare$} & \textcolor{nbicycle}{$\blacksquare$} & \textcolor{nbus}{$\blacksquare$} & \textcolor{ncar}{$\blacksquare$} & \textcolor{nconstruct}{$\blacksquare$} & \textcolor{nmotor}{$\blacksquare$} & \textcolor{npedestrian}{$\blacksquare$} & \textcolor{ntraffic}{$\blacksquare$} & \textcolor{ntrailer}{$\blacksquare$} & \textcolor{ntruck}{$\blacksquare$} & \textcolor{ndriveable}{$\blacksquare$} & \textcolor{nother}{$\blacksquare$} & \textcolor{nsidewalk}{$\blacksquare$} & \textcolor{nterrain}{$\blacksquare$} & \textcolor{nmanmade}{$\blacksquare$} & 
            \textcolor{nvegetation}{$\blacksquare$} \\
            barrier & bicycle & bus & car & const. veh. & motorcycle & pedestrian & traffic cone & trailer & truck & drive. surf. & other flat & sidewalk & terrain & manmade & vegetation
        \end{tabular}
    \vspace{2pt}

    \captionsetup[subfigure]{labelformat=empty}
    \centering
    \begin{subfigure}[b]{0.45\textwidth}
        \centering
        \includegraphics[width=\textwidth]{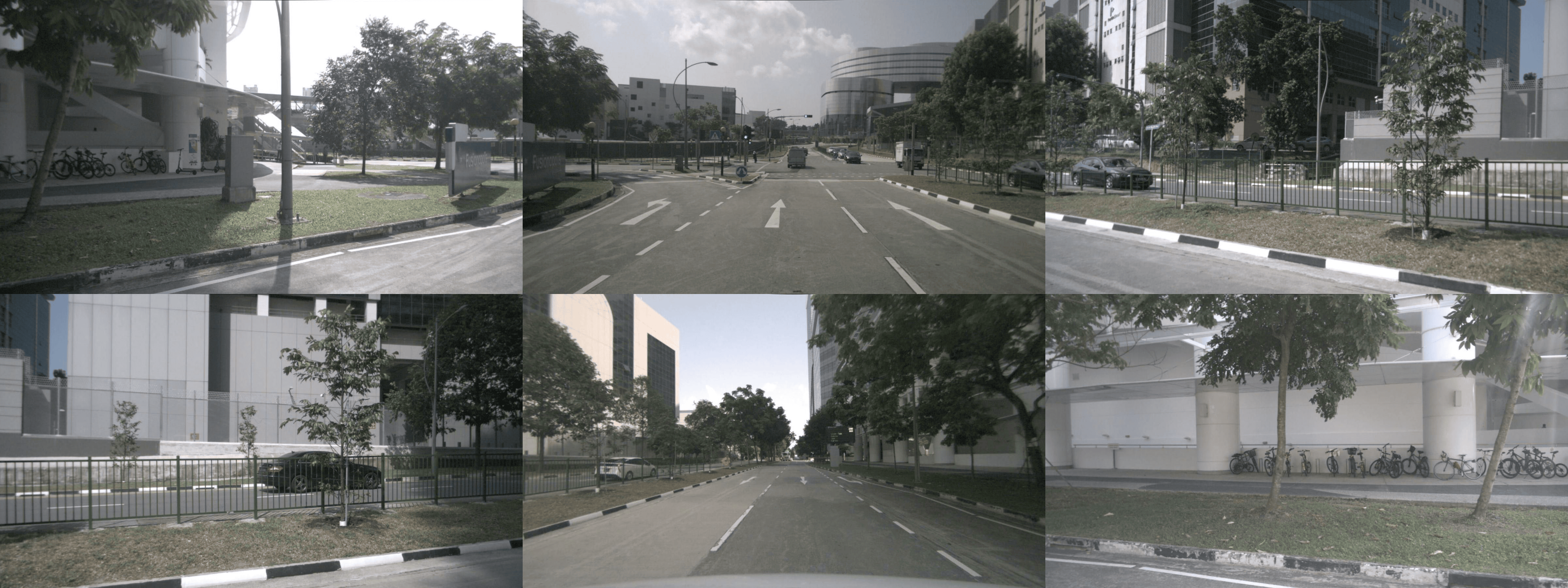}
    \end{subfigure}
    \hspace{1pt}
    \begin{subfigure}[b]{0.17\textwidth}
        \centering
        \includegraphics[width=\textwidth]{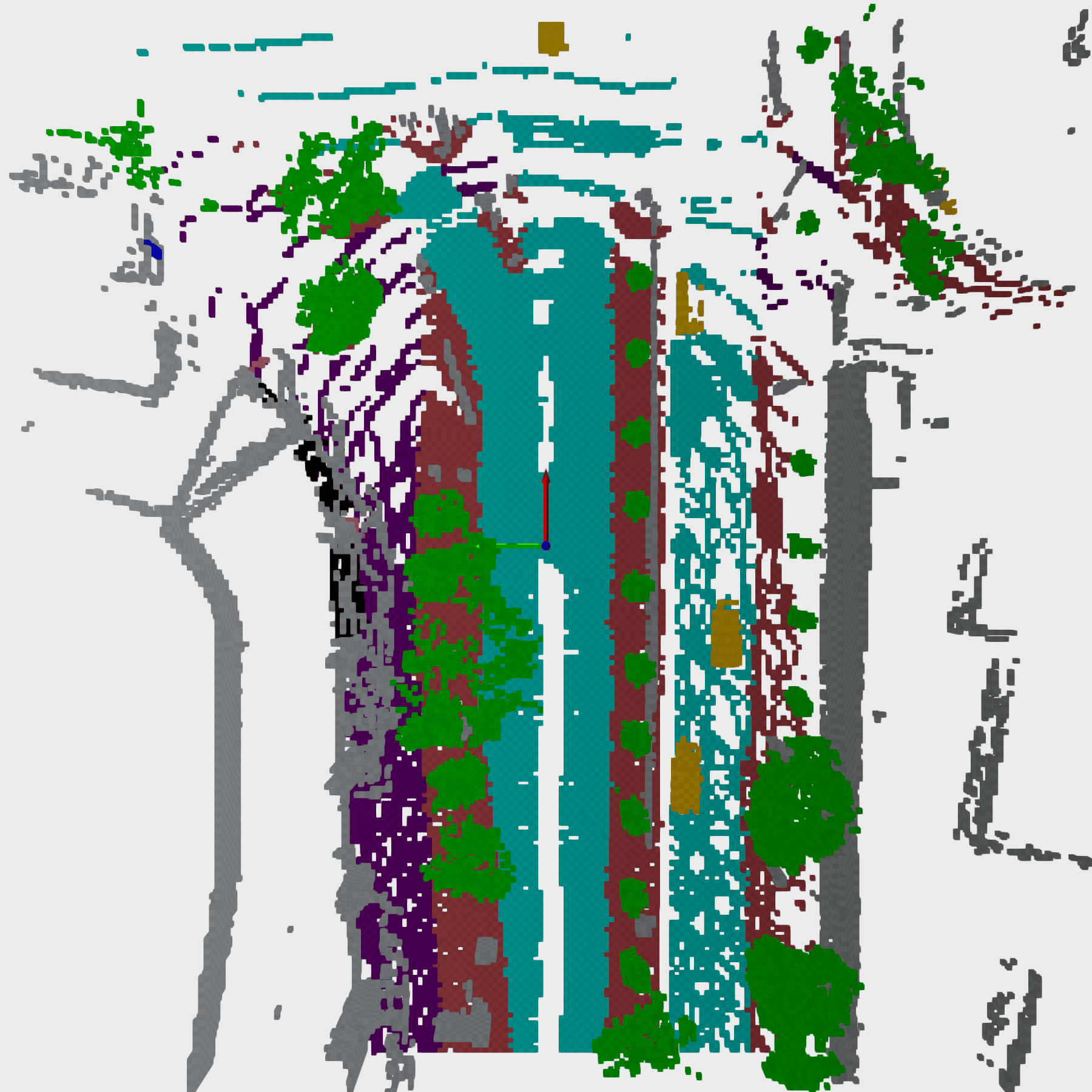}
    \end{subfigure}
    \hspace{1pt}
    \begin{subfigure}[b]{0.17\textwidth}
        \centering
        \includegraphics[width=\textwidth]{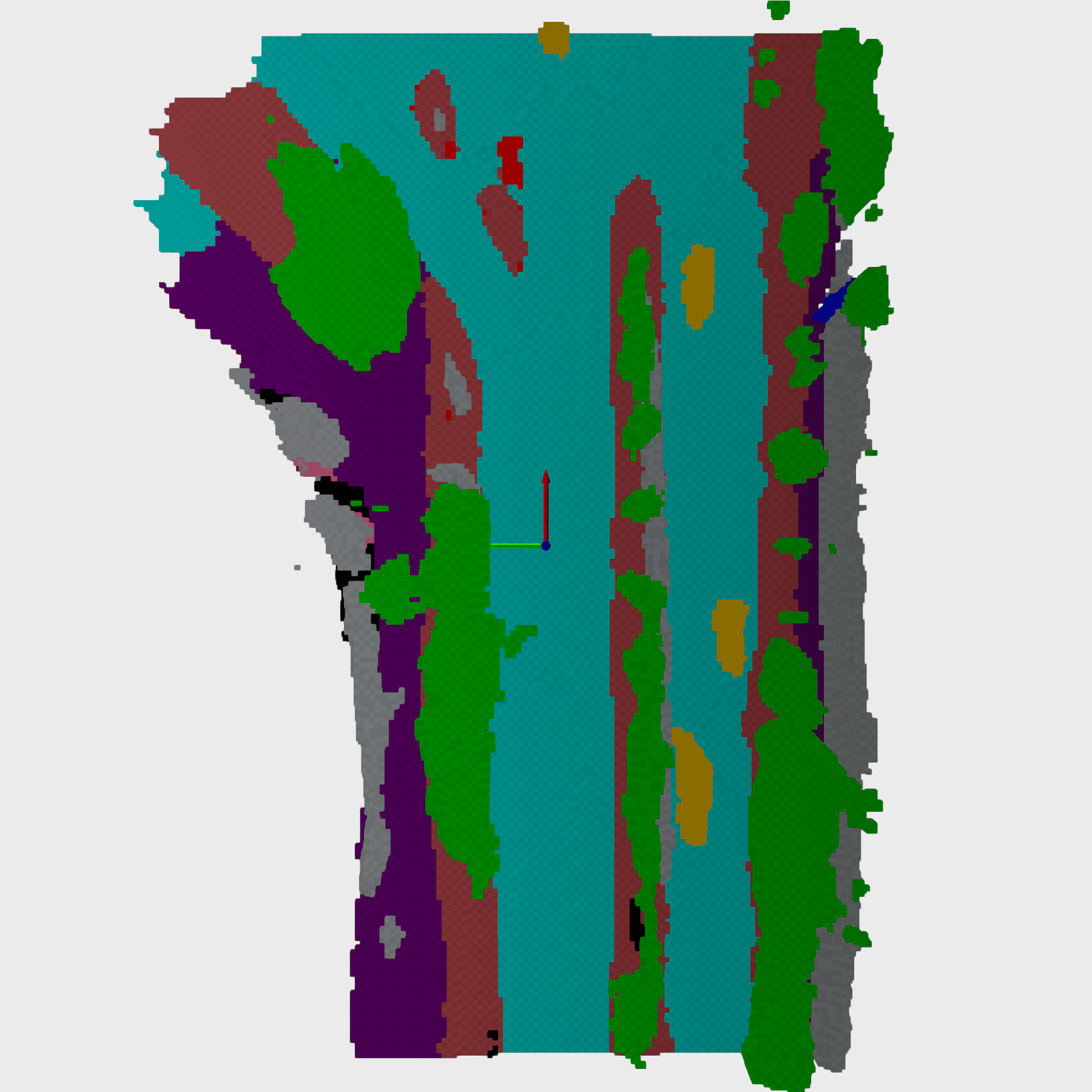}
    \end{subfigure}
    \hspace{1pt}
    \begin{subfigure}[b]{0.17\textwidth}
        \centering
        \includegraphics[width=\textwidth]{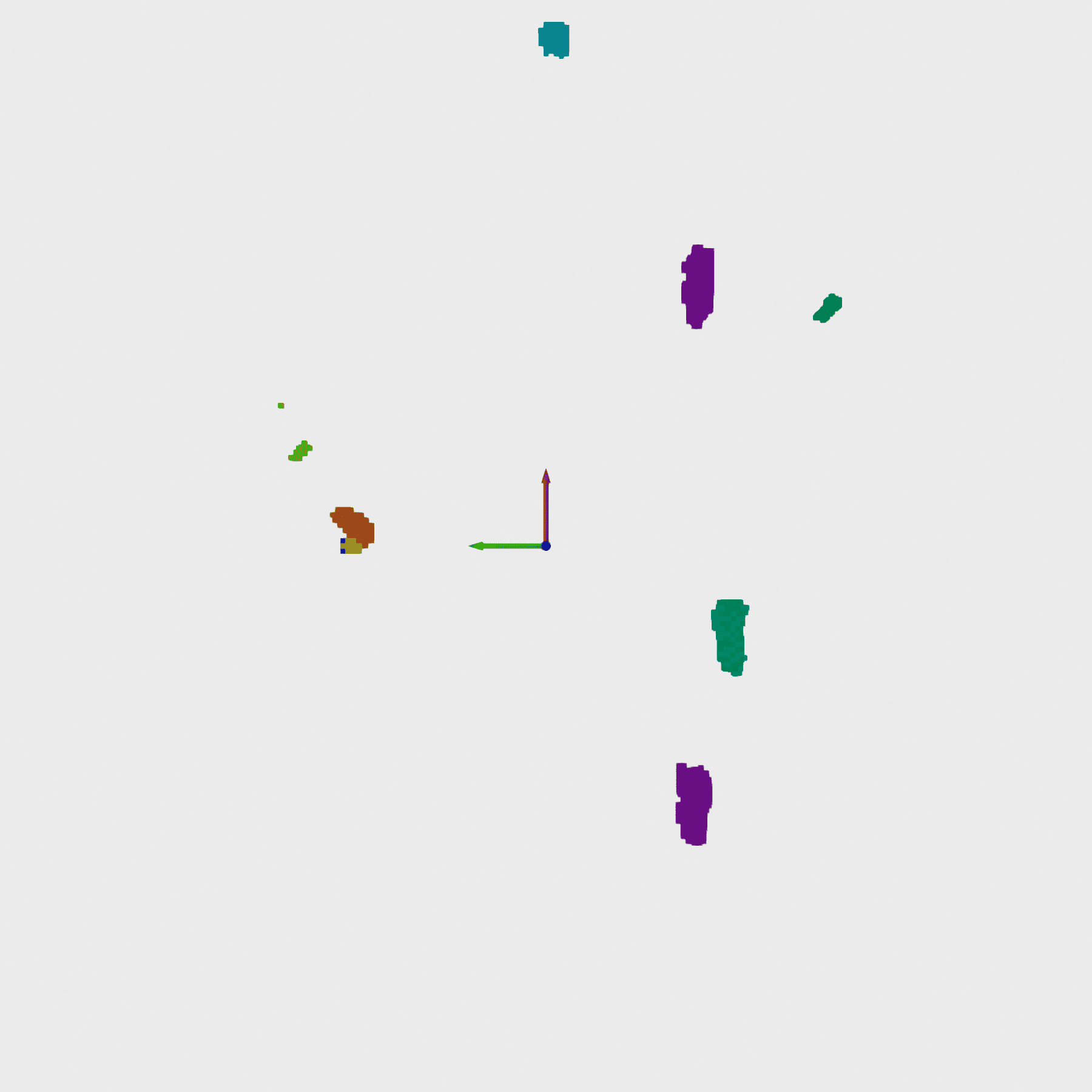}
    \end{subfigure}

    \vspace{3pt}
    
    \begin{subfigure}[b]{0.45\textwidth}
        \centering
        \includegraphics[width=\textwidth]{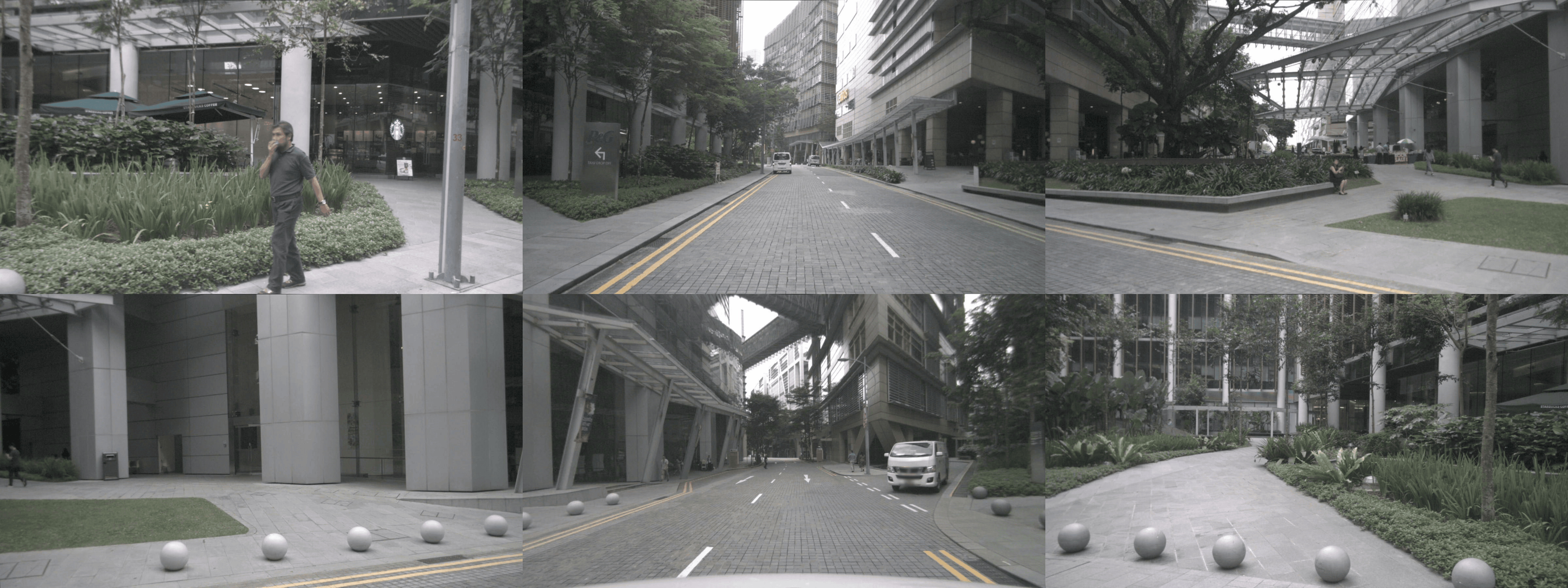}
    \end{subfigure}
    \hspace{1pt}
    \begin{subfigure}[b]{0.17\textwidth}
        \centering
        \includegraphics[width=\textwidth]{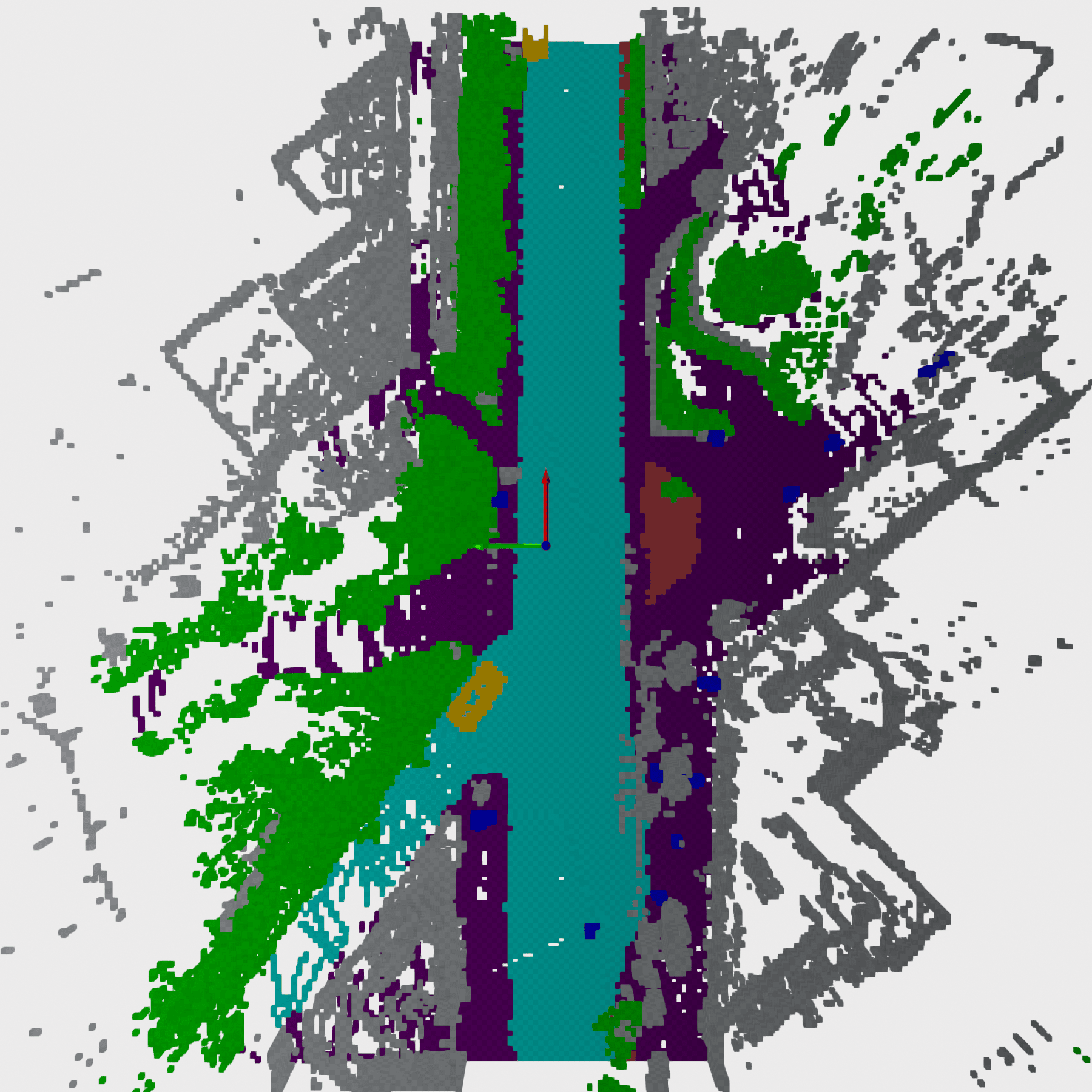}
    \end{subfigure}
    \hspace{1pt}
    \begin{subfigure}[b]{0.17\textwidth}
        \centering
        \includegraphics[width=\textwidth]{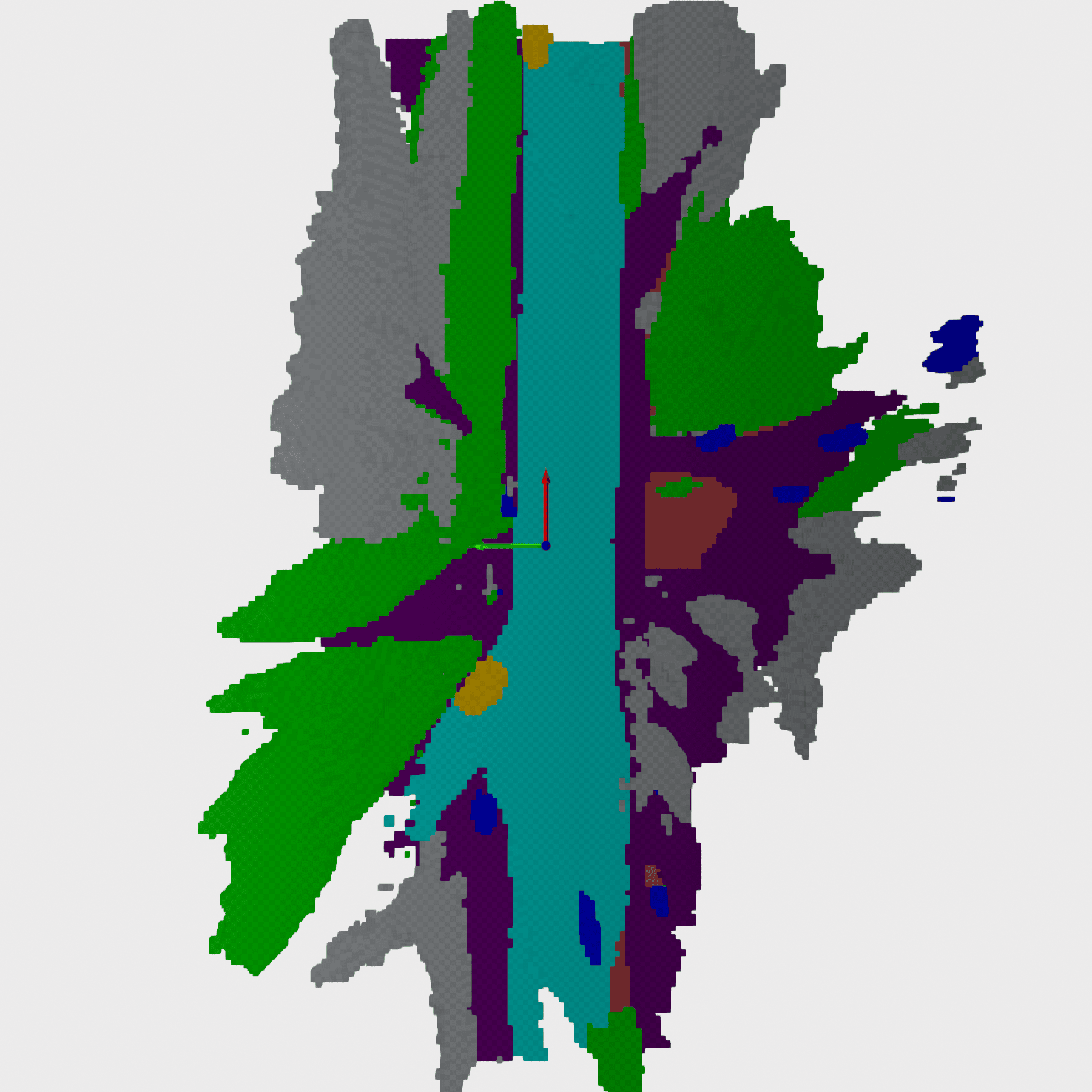}
    \end{subfigure}
    \hspace{1pt}
    \begin{subfigure}[b]{0.17\textwidth}
        \centering
        \includegraphics[width=\textwidth]{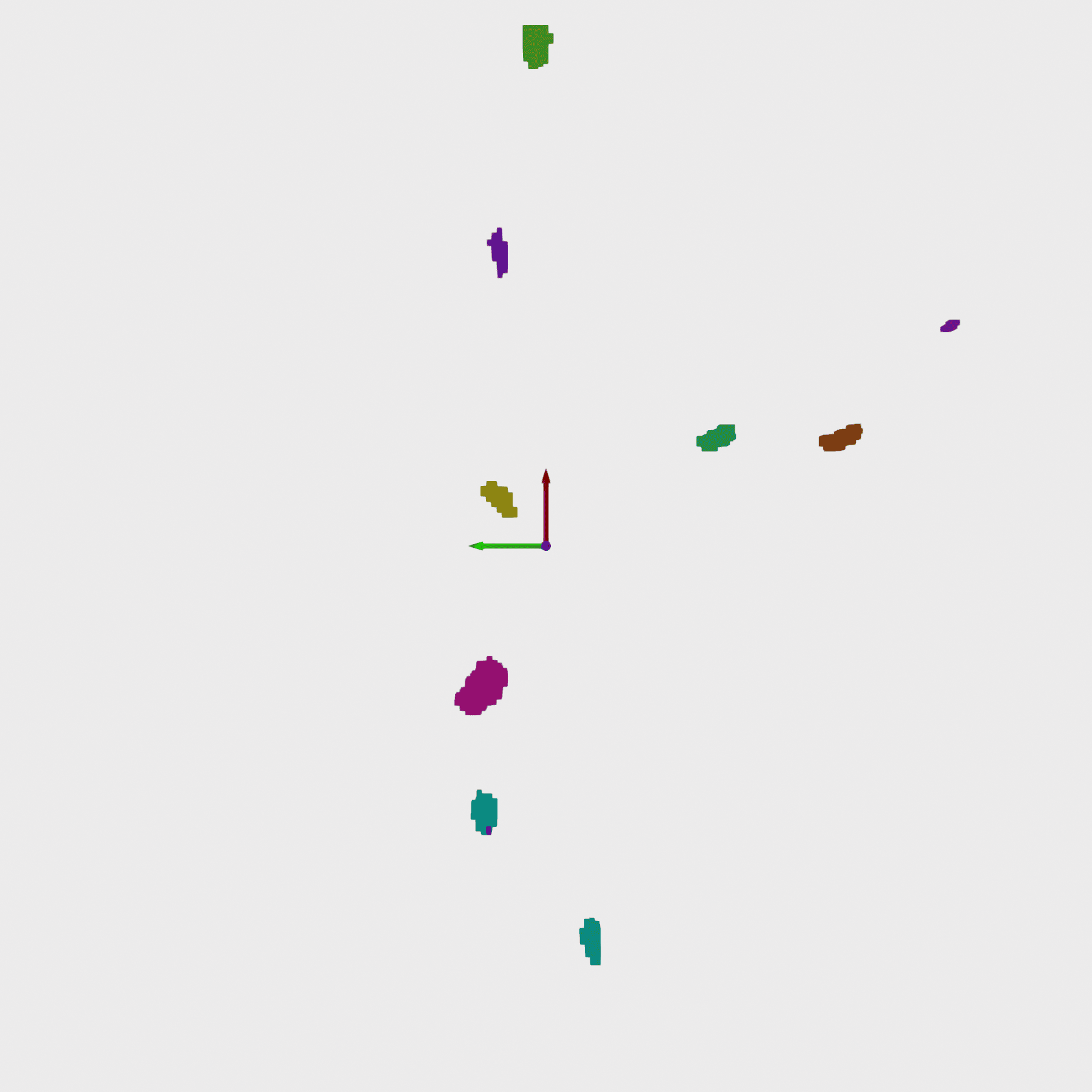}
    \end{subfigure}

    \vspace{3pt}
    
    \begin{subfigure}[b]{0.45\textwidth}
        \centering
        \includegraphics[width=\textwidth]{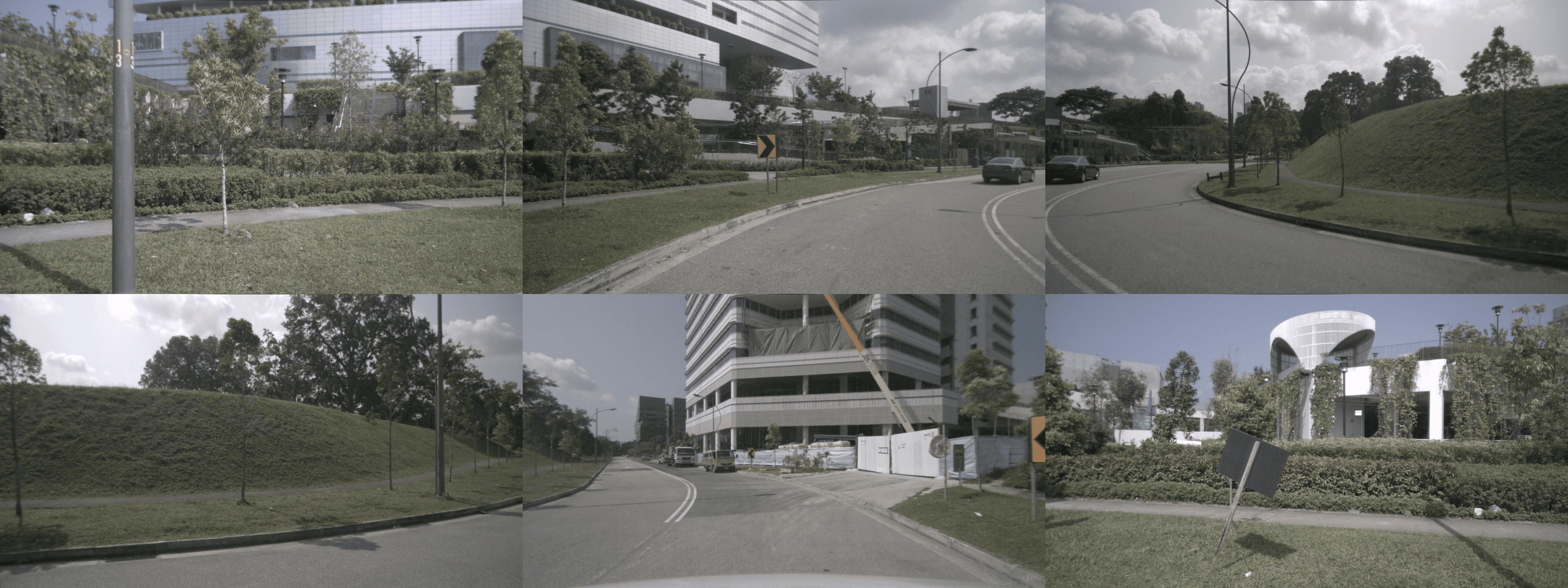}
        \caption{\textbf{Multi-View Images}}
    \end{subfigure}
    \hspace{1pt}
    \begin{subfigure}[b]{0.17\textwidth}
        \centering
        \includegraphics[width=\textwidth]{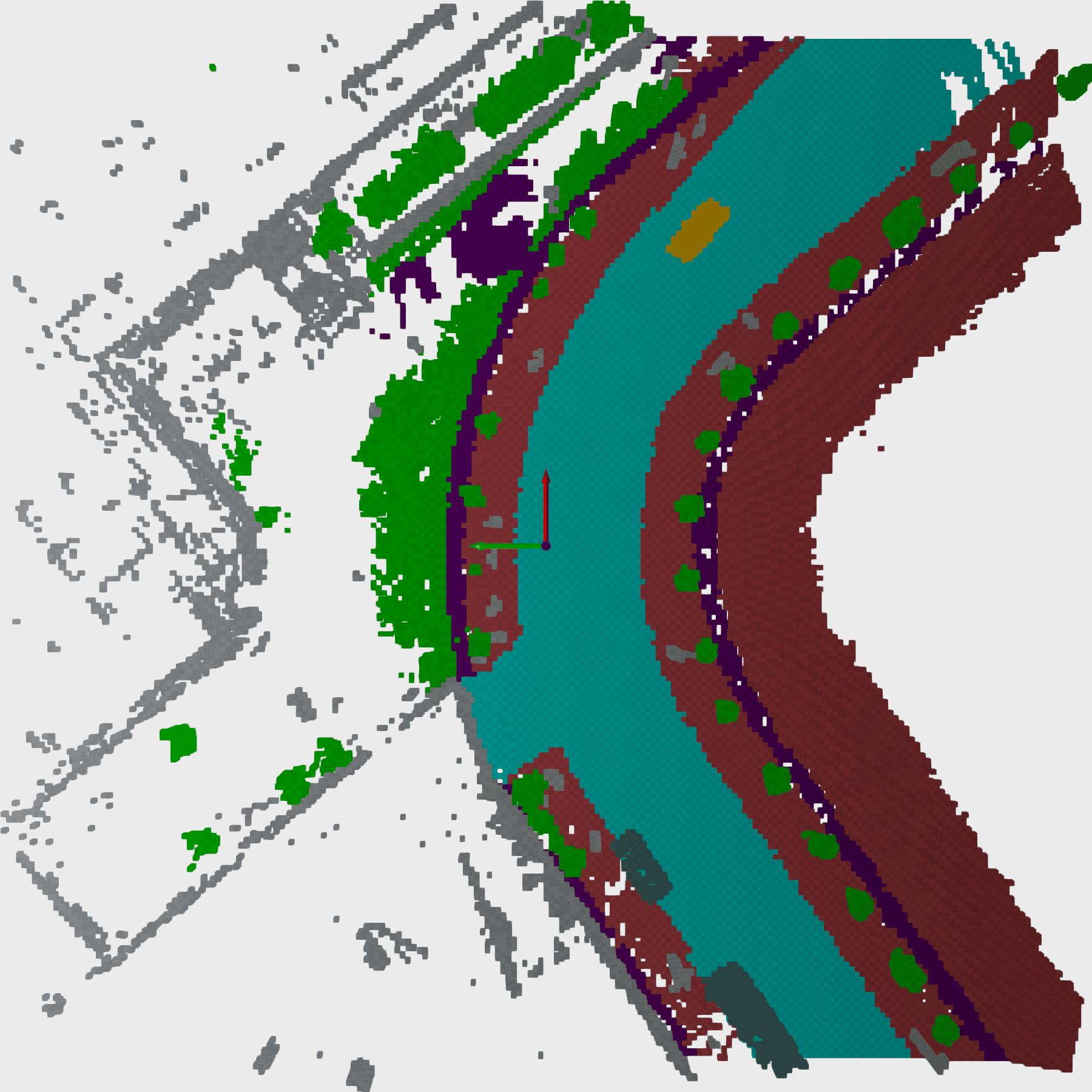}
        \caption{\textbf{GT occupancy}}
    \end{subfigure}
    \hspace{1pt}
    \begin{subfigure}[b]{0.17\textwidth}
        \centering
        \includegraphics[width=\textwidth]{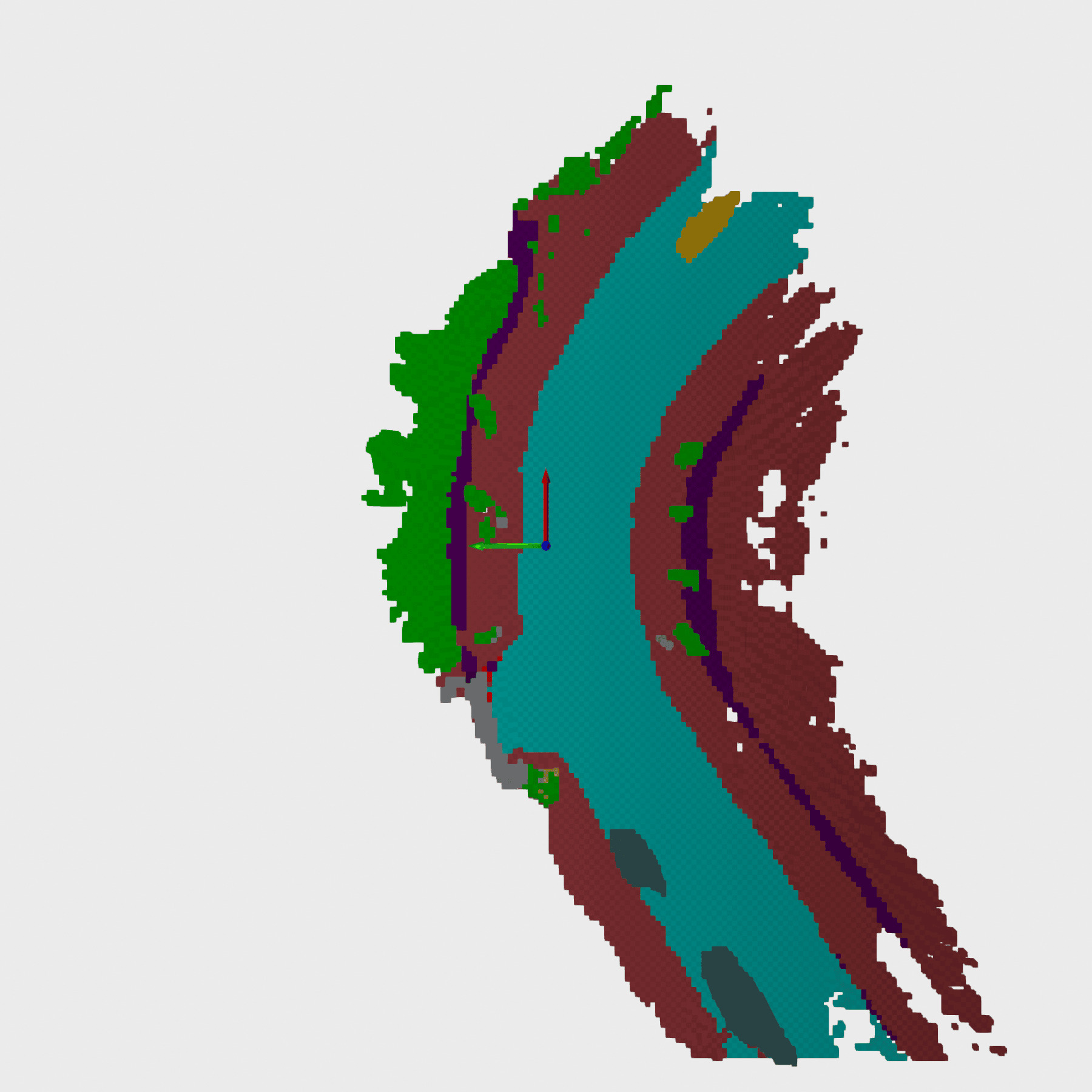}
        \caption{\textbf{Pred occupancy}}
    \end{subfigure}
    \hspace{1pt}
    \begin{subfigure}[b]{0.17\textwidth}
        \centering
        \includegraphics[width=\textwidth]{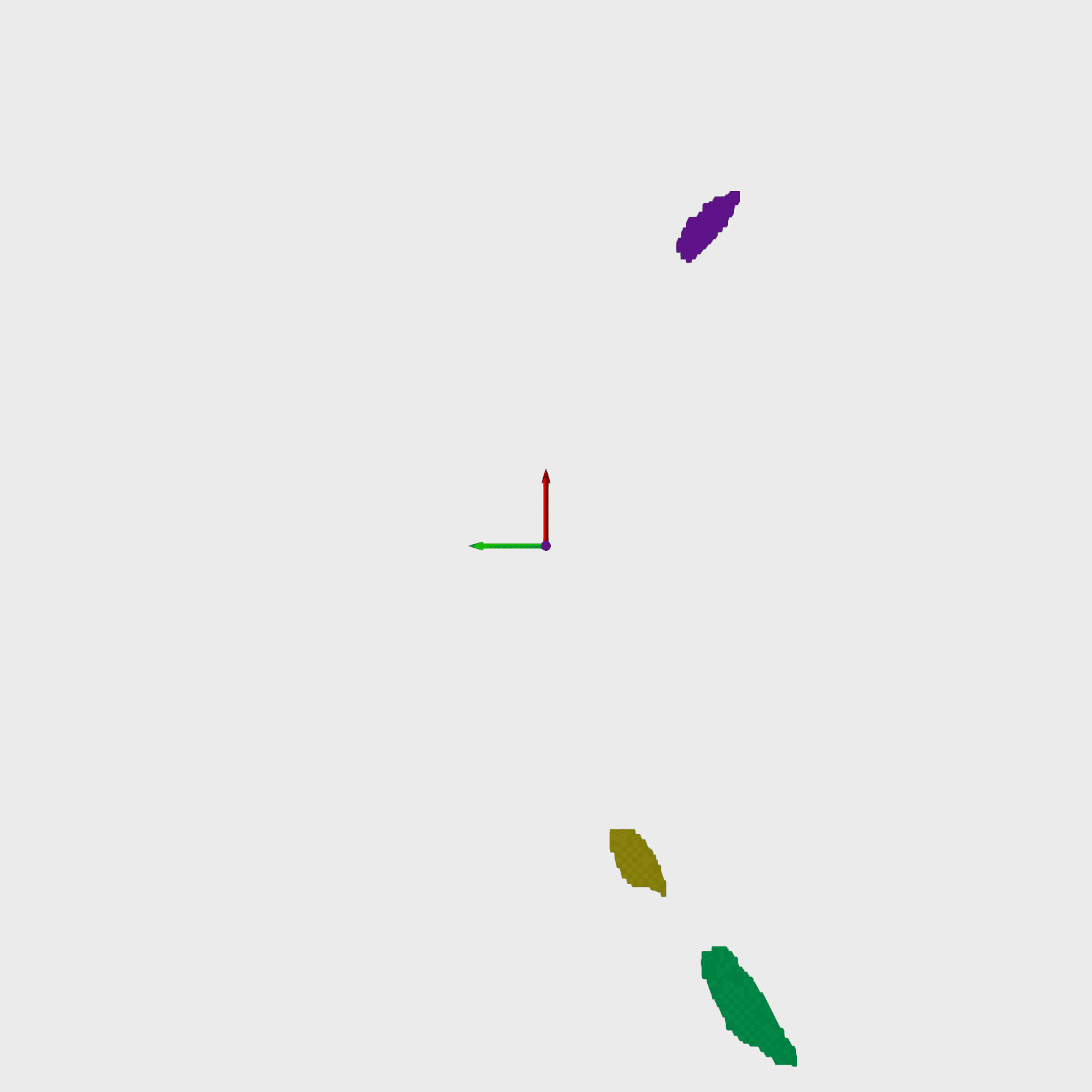}
        \caption{\textbf{Pred panoptic}}
    \end{subfigure}

    \end{minipage}
    
    \caption{Additional qualitative results of our model \ourmodelname{}.}
    
    \label{fig:qualitative_examples_supp}
\end{figure*}

%% file: table/nuscenes_occ_val.tex
\input{misc/occupancy_colors}
\begin{table*}[!h]
	\footnotesize
 	\setlength{\tabcolsep}{0.005\linewidth}
	
	\newcommand{\classfreq}[1]{{~\tiny(\nuscenesfreq{#1}\%)}}  %
    \def\b{\textbf}
    \def\u{\underline}
    
    \begin{center}

	\begin{tabular}{l | c c c c | r r | r r r r r r r r r}
		\toprule
		Method & \makecell{Image \\ Backbone} & \makecell{Temporal} & \makecell{Train \\ w/ mask} & \makecell{Evaluate \\ w/ mask} & mIoU & IoU
		
		& \rotatebox{90}{\textcolor{nbicycle}{$\blacksquare$} bicycle}
		
		& \rotatebox{90}{\textcolor{nbus}{$\blacksquare$} bus}

		& \rotatebox{90}{\textcolor{ncar}{$\blacksquare$} car}

		& \rotatebox{90}{\textcolor{nconstruct}{$\blacksquare$} const. veh.}

		& \rotatebox{90}{\textcolor{nmotor}{$\blacksquare$} motorcycle}

		& \rotatebox{90}{\textcolor{npedestrian}{$\blacksquare$} pedestrian}

		& \rotatebox{90}{\textcolor{ntraffic}{$\blacksquare$} traffic cone}

		& \rotatebox{90}{\textcolor{ntrailer}{$\blacksquare$} trailer}

		& \rotatebox{90}{\textcolor{ntruck}{$\blacksquare$} truck}

		\\
		\midrule
        MonoScene~\cite{Cao2022}                & R101-DCN  & \xmark & \xmark & \cmark &     6.1  &       -  &     4.3  &     4.9  &     9.4  &     5.7  &     4.0  &     3.0  &     5.9  &     4.4  &  7.2 \\
        BEVDet~\cite{Huang2021}                 & R101-DCN  & \xmark & \xmark & \cmark &    19.4  &       -  &     0.2  &    32.3  &    34.5  &    13.0  &    10.3  &    10.4  &     6.3  &     8.9  & 23.6 \\
        OccFormer \cite{Zhang2023}              & R101      & \xmark & \xmark & \cmark &    21.9  &       -  &    12.3  &    34.4  &    39.2  &    14.4  &    16.4  &    17.2  &     9.3  &    13.9  & 26.4 \\
        BEVFormer~\cite{Li2022}                 & R101-DCN  & \cmark & \xmark & \cmark &    26.9  &       -  &    17.9  &    40.4  &    42.4  &     7.4  &    23.9  &    21.8  &    21.0  &    22.4  & 30.7 \\
        TPVFormer~\cite{Huang2023}              & R101-DCN  & \cmark & \xmark & \cmark &    27.8  &       -  &    13.7  &    40.8  &    45.9  &    17.2  &    20.0  &    18.8  &    14.3  &    26.7  & 34.2 \\
        CTF-Occ~\cite{Tian2023}                 & R101-DCN  & \xmark & \xmark & \cmark &    28.5  &       -  &    20.6  &    38.3  &    42.2  &    16.9  &    24.5  &    22.7  &    21.0  &    23.0  & 31.1 \\
        SparseOcc~\cite{Liu2023}                & R50       & \cmark & \xmark & \cmark &    30.9  &       -  &       -  &       -  &       -  &       -  &       -  &       -  &       -  &       -  &    - \\
        PanoOcc~\cite{Wang2024PanoOcc}          & R101-DCN  & \cmark & \xmark & \cmark &    32.5  &       -  & \u{27.2} &    43.5  &    48.7  & \u{23.0} & \u{31.2} &    27.6  & \u{28.6} &    26.6  & 38.3 \\
        \midrule
        TPVFormer$\ddagger$~\cite{Huang2023}    & R50       & \cmark & \cmark & \cmark &    34.2  &    66.8  &    17.7  &    40.9  &    47.0  &    15.1  &    20.5  &    24.7  &    24.7  &    24.3  & 29.3 \\
        OccFormer$\ddagger$~\cite{Zhang2023}    & R50       & \xmark & \cmark & \cmark &    37.4  & \u{70.1} &    18.2  &    42.8  &    50.3  &    24.0  &    20.8  &    22.9  &    21.0  &    31.9  & 38.1 \\ 
        BEVFormer~\cite{Li2022}                 & R101-DCN  & \cmark & \cmark & \cmark & \u{39.2} &       -  &    24.9  & \u{47.6} & \u{54.5} &    20.2  &    28.8  & \u{28.0} &    25.7  & \u{33.0} & \u{38.6} \\
        PanoOcc~\cite{Wang2024PanoOcc}          & R101-DCN  & \cmark & \cmark & \cmark & \b{44.5} & \b{75.0} & \b{29.6} & \b{49.4} & \b{55.5} & \b{23.3} & \b{33.3} & \b{30.6} & \b{31.0} & \b{34.4} & \b{42.6} \\
        \midrule
        \textbf{\ourmodelname{} (Ours)}         & R101      & \xmark & \xmark & \cmark &    28.0  &    43.9  &    21.6  &    39.0  &    43.3  &    18.3  &    21.8  &    20.2  &    14.2  &    19.9  & 30.3 \\
        \textbf{\ourmodelname{} (Ours)}         & R101      & \xmark & \xmark & \xmark &    17.2  &    24.9  &     2.3  &    27.9  &    30.6  &    11.2  &    12.2  &    13.3  &     6.2  &    10.6  & 21.3 \\
	\bottomrule
	\end{tabular}
    \end{center}
    \vspace{-12pt}
    \caption{\textbf{3\textsc{d} Occupancy prediction performance on the Occ3D-nuScenes dataset \emph{things} classes.} \say{Temporal} indicates that the model uses past frames when generating predictions. \say{Train w/ mask} and \say{Evaluate w/ mask} indicate whether the model has been trained using the camera mask and whether the performance has been measured using the camera mask, respectively. $\ddagger$ indicates performance measured by \cite{Ma2024}. Best performance is \textbf{bolded} and second best is \underline{underlined}.}
    \label{tab:camera_occ_things}
\end{table*}

\begin{table*}[ht]
	\footnotesize
 	\setlength{\tabcolsep}{0.005\linewidth}
	
	\newcommand{\classfreq}[1]{{~\tiny(\nuscenesfreq{#1}\%)}}  %
    \def\b{\textbf}
    \def\u{\underline}
    
    \begin{center}

	\begin{tabular}{l | c c c c | r r | r r r r r r r r}
		\toprule
		Method & \makecell{Image \\ Backbone} & \makecell{Temporal} & \makecell{Train \\ w/ mask} & \makecell{Evaluate \\ w/ mask} & mIoU & IoU
        & \rotatebox{90}{\textcolor{nothers}{$\blacksquare$} others}
        
		& \rotatebox{90}{\textcolor{nbarrier}{$\blacksquare$} barrier}
		
		& \rotatebox{90}{\textcolor{ndriveable}{$\blacksquare$} drive. suf.}

		& \rotatebox{90}{\textcolor{nother}{$\blacksquare$} other flat}

		& \rotatebox{90}{\textcolor{nsidewalk}{$\blacksquare$} sidewalk}

		& \rotatebox{90}{\textcolor{nterrain}{$\blacksquare$} terrain}

		& \rotatebox{90}{\textcolor{nmanmade}{$\blacksquare$} manmade}

		& \rotatebox{90}{\textcolor{nvegetation}{$\blacksquare$} vegetation}

		\\
		\midrule
        MonoScene~\cite{Cao2022}                & R101-DCN  & \xmark & \xmark & \cmark &     6.1  &       -  &     1.8  &     7.2  &    14.9  &     6.3  &     7.9  &     7.4  &     1.0  &     7.6  \\
        BEVDet~\cite{Huang2021}                 & R101-DCN  & \xmark & \xmark & \cmark &    19.4  &       -  &     4.4  &    30.3  &    52.3  &    24.6  &    26.1  &    22.3  &    15.0  &    15.1  \\
        OccFormer \cite{Zhang2023}              & R101      & \xmark & \xmark & \cmark &    21.9  &       -  &     5.9  &    30.3  &    51.0  &    31.0  &    34.7  &    22.7  &     6.8  &     7.0  \\
        BEVFormer~\cite{Li2022}                 & R101-DCN  & \cmark & \xmark & \cmark &    26.9  &       -  &     5.8  &    37.8  &    55.4  &    28.4  &    36.0  &    28.1  &    20.0  &    17.7  \\
        TPVFormer~\cite{Huang2023}              & R101-DCN  & \cmark & \xmark & \cmark &    27.8  &       -  &     7.2  &    38.9  &    55.6  &    35.5  &    37.6  &    30.7  &    19.4  &    16.8  \\
        CTF-Occ~\cite{Tian2023}                 & R101-DCN  & \xmark & \xmark & \cmark &    28.5  &       -  &     8.1  &    39.3  &    53.3  &    33.8  &    38.0  &    33.2  &    20.8  &    18.0  \\
        SparseOcc~\cite{Liu2023}                & R50       & \cmark & \xmark & \cmark &    30.9  &       -  &       -  &       -  &       -  &       -  &       -  &       -  &       -  &       -  \\
        PanoOcc~\cite{Wang2024PanoOcc}          & R101-DCN  & \cmark & \xmark & \cmark &    32.5  &       -  & \u{10.8} &    46.9  &    58.0  &    38.9  &    38.2  &    32.3  &    15.6  &    16.4  \\
        \midrule
        TPVFormer$\ddagger$~\cite{Huang2023}    & R50       & \cmark & \cmark & \cmark &    34.2  &    66.8  &     7.7  &    44.0  &    79.3  &    40.6  &    48.5  &    49.4  &    32.6  &    29.8  \\
        OccFormer$\ddagger$~\cite{Zhang2023}    & R50       & \xmark & \cmark & \cmark &    37.4  & \u{70.1} &     9.2  &    45.8  &    80.1  &    38.2  &    50.8  & \u{54.3} & \b{46.4} & \u{40.2} \\ 
        BEVFormer~\cite{Li2022}                 & R101-DCN  & \cmark & \cmark & \cmark & \u{39.2} &       -  &    10.1  & \u{47.9} & \u{82.0} & \u{40.6} & \u{50.9} &    53.0  &    43.9  &    37.2  \\
        PanoOcc~\cite{Wang2024PanoOcc}          & R101-DCN  & \cmark & \cmark & \cmark & \b{44.5} & \b{75.0} & \b{11.7} & \b{50.5} & \b{83.3} & \b{44.2} & \b{54.4} & \b{56.0} & \u{45.9} & \b{40.4} \\
        \midrule
        \textbf{\ourmodelname{} (Ours)}         & R101      & \xmark & \xmark & \cmark &    28.0  &    43.9  &     3.7  &    35.7  &    61.2  &    30.5  &    38.1  &    36.4  &    19.3  &    22.6  \\
        \textbf{\ourmodelname{} (Ours)}         & R101      & \xmark & \xmark & \xmark &    17.2  &    24.9  &     2.3  &    19.5  &    36.9  &    20.1  &    22.9  &    17.8  &    11.3  &    16.8  \\
	\bottomrule
	\end{tabular}
    \end{center}
    \vspace{-12pt}
    \caption{\textbf{3\textsc{d} Occupancy prediction performance on the Occ3D-nuScenes dataset \emph{stuff} classes.} \say{Temporal} indicates that the model uses past frames when generating predictions. \say{Train w/ mask} and \say{Evaluate w/ mask} indicate whether the model has been trained using the camera mask and whether the performance has been measured using the camera mask, respectively. $\ddagger$ indicates performance measured by \cite{Ma2024}. Best performance is \textbf{bolded} and second best is \underline{underlined}.}
    \label{tab:camera_occ_stuff}
\end{table*}

%% file: main.bbl
\begin{thebibliography}{51}
\providecommand{\natexlab}[1]{#1}
\providecommand{\url}[1]{\texttt{#1}}
\expandafter\ifx\csname urlstyle\endcsname\relax
  \providecommand{\doi}[1]{doi: #1}\else
  \providecommand{\doi}{doi: \begingroup \urlstyle{rm}\Url}\fi

\bibitem[Caesar et~al.(2019)Caesar, Bankiti, Lang, Vora, Liong, Xu, Krishnan,
  Pan, Baldan, and Beijbom]{nuscenes2019}
Holger Caesar, Varun Bankiti, Alex~H. Lang, Sourabh Vora, Venice~Erin Liong,
  Qiang Xu, Anush Krishnan, Yu Pan, Giancarlo Baldan, and Oscar Beijbom.
\newblock nuscenes: A multimodal dataset for autonomous driving.
\newblock \emph{arXiv preprint arXiv:1903.11027}, 2019.

\bibitem[Cao and de~Charette(2022)]{Cao2022}
Anh-Quan Cao and Raoul de Charette.
\newblock Monoscene: Monocular 3d semantic scene completion.
\newblock In \emph{IEEE/CVF Conference on Computer Vision and Pattern
  Recognition (CVPR)}, pages 3991--4001, 2022.

\bibitem[Carion et~al.(2020)Carion, Massa, Synnaeve, Usunier, Kirillov, and
  Zagoruyko]{carion2020end}
Nicolas Carion, Francisco Massa, Gabriel Synnaeve, Nicolas Usunier, Alexander
  Kirillov, and Sergey Zagoruyko.
\newblock End-to-end object detection with transformers.
\newblock In \emph{European conference on computer vision}, pages 213--229.
  Springer, 2020.

\bibitem[Chen et~al.(2020)Chen, Liu, Zhao, Wang, and Jia]{chen2020gridmask}
Pengguang Chen, Shu Liu, Hengshuang Zhao, Xingquan Wang, and Jiaya Jia.
\newblock Gridmask data augmentation.
\newblock \emph{arXiv preprint arXiv:2001.04086}, 2020.

\bibitem[Deng et~al.(2009)Deng, Dong, Socher, Li, Li, and
  Li]{conf-cvpr-DengDSLL009}
Jia Deng, Wei Dong, Richard Socher, Li-Jia Li, Kai Li, and Fei-Fei Li.
\newblock Imagenet: A large-scale hierarchical image database.
\newblock In \emph{CVPR}, pages 248--255. IEEE Computer Society, 2009.

\bibitem[Gu and Dao(2023)]{Gu2023}
Albert Gu and Tri Dao.
\newblock Mamba: Linear-time sequence modeling with selective state spaces.
\newblock \emph{arXiv preprint}, 2023.

\bibitem[He et~al.(2016)He, Zhang, Ren, and Sun]{he2016deepResNet}
Kaiming He, Xiangyu Zhang, Shaoqing Ren, and Jian Sun.
\newblock Deep residual learning for image recognition.
\newblock In \emph{Proceedings of the IEEE conference on computer vision and
  pattern recognition}, pages 770--778, 2016.

\bibitem[He et~al.(2024)He, Chen, Xun, and Tan]{He2024}
Yulin He, Wei Chen, Tianci Xun, and Yusong Tan.
\newblock Real-time 3d occupancy prediction via geometric-semantic
  disentanglement.
\newblock \emph{arXiv}, 2024.

\bibitem[Huang et~al.(2021)Huang, Huang, Zhu, Ye, and Du]{Huang2021}
Junjie Huang, Guan Huang, Zheng Zhu, Yun Ye, and Dalong Du.
\newblock Bevdet: High-performance multi-camera 3d object detection in
  bird-eye-view.
\newblock \emph{arXiv}, 2021.

\bibitem[Huang et~al.(2023)Huang, Zheng, Zhang, Zhou, and Lu]{Huang2023}
Yuanhui Huang, Wenzhao Zheng, Yunpeng Zhang, Jie Zhou, and Jiwen Lu.
\newblock Tri-perspective view for vision-based 3d semantic occupancy
  prediction.
\newblock In \emph{IEEE/CVF Conference on Computer Vision and Pattern
  Recognition (CVPR)}, pages 9223--9232, 2023.

\bibitem[Huang et~al.(2024{\natexlab{a}})Huang, Zheng, Zhang, Zhou, and
  Lu]{Huang2024Self}
Yuanhui Huang, Wenzhao Zheng, Borui Zhang, Jie Zhou, and Jiwen Lu.
\newblock Selfocc: Self-supervised vision-based 3d occupancy prediction.
\newblock In \emph{IEEE/CVF Conference on Computer Vision and Pattern
  Recognition (CVPR)}, pages 19946--19956, 2024{\natexlab{a}}.

\bibitem[Huang et~al.(2024{\natexlab{b}})Huang, Zheng, Zhang, Zhou, and
  Lu]{Huang2024}
Yuanhui Huang, Wenzhao Zheng, Yunpeng Zhang, Jie Zhou, and Jiwen Lu.
\newblock Gaussianformer: Scene as gaussians for vision-based 3d semantic
  occupancy prediction.
\newblock \emph{arXiv}, 2024{\natexlab{b}}.

\bibitem[Kazhdan et~al.(2006)Kazhdan, Bolitho, and Hoppe]{Kazhdan2006}
M Kazhdan, M Bolitho, and H Hoppe.
\newblock Poisson surface reconstruction.
\newblock In \emph{Proceedings of the fourth Eurographics symposium on Geometry
  processing}, 2006.

\bibitem[Kirillov et~al.(2019)Kirillov, He, Girshick, Rother, and
  Doll{\'a}r]{kirillov2019panoptic}
Alexander Kirillov, Kaiming He, Ross Girshick, Carsten Rother, and Piotr
  Doll{\'a}r.
\newblock Panoptic segmentation.
\newblock In \emph{Proceedings of the IEEE/CVF conference on computer vision
  and pattern recognition}, pages 9404--9413, 2019.

\bibitem[Kuhn(1955)]{kuhn1955hungarian}
Harold~W Kuhn.
\newblock The hungarian method for the assignment problem.
\newblock \emph{Naval research logistics quarterly}, 2\penalty0 (1-2):\penalty0
  83--97, 1955.

\bibitem[Lecun et~al.(2015)Lecun, Bengio, and Hinton]{Lecun2015}
Yann Lecun, Yoshua Bengio, and Geoffrey Hinton.
\newblock Deep learning.
\newblock \emph{Nature 2015 521:7553}, 521:\penalty0 436--444, 2015.

\bibitem[Li et~al.(2024)Li, Hou, Xing, Sun, and Zhang]{Li2024}
Heng Li, Yuenan Hou, Xiaohan Xing, Xiao Sun, and Yanyong Zhang.
\newblock Occmamba: Semantic occupancy prediction with state space models.
\newblock \emph{arXiv}, 2024.

\bibitem[Li et~al.(2022{\natexlab{a}})Li, He, Wen, Gao, Cheng, and
  Zhang]{li2022panoptic}
Jinke Li, Xiao He, Yang Wen, Yuan Gao, Xiaoqiang Cheng, and Dan Zhang.
\newblock Panoptic-phnet: Towards real-time and high-precision lidar panoptic
  segmentation via clustering pseudo heatmap.
\newblock In \emph{Proceedings of the IEEE/CVF Conference on Computer Vision
  and Pattern Recognition}, pages 11809--11818, 2022{\natexlab{a}}.

\bibitem[Li et~al.(2023{\natexlab{a}})Li, Yu, Choy, Xiao, Alvarez, Fidler,
  Feng, and Anandkumar]{Li2023VoxFormer}
Yiming Li, Zhiding Yu, Christopher Choy, Chaowei Xiao, Jose~M. Alvarez, Sanja
  Fidler, Chen Feng, and Anima Anandkumar.
\newblock Voxformer: Sparse voxel transformer for camera-based 3d semantic
  scene completion.
\newblock pages 9087--9098, 2023{\natexlab{a}}.

\bibitem[Li et~al.(2022{\natexlab{b}})Li, Wang, Li, Xie, Sima, Lu, Qiao, and
  Dai]{Li2022}
Zhiqi Li, Wenhai Wang, Hongyang Li, Enze Xie, Chonghao Sima, Tong Lu, Yu Qiao,
  and Jifeng Dai.
\newblock Bevformer: Learning bird’s-eye-view representation
  from multi-camera images via spatiotemporal transformers.
\newblock \emph{ECCV 2022. Lecture Notes in Computer Science}, 13669:\penalty0
  1--18, 2022{\natexlab{b}}.

\bibitem[Li et~al.(2023{\natexlab{b}})Li, Yu, Austin, Fang, Lan, Kautz, and
  Alvarez]{Li2023}
Zhiqi Li, Zhiding Yu, David Austin, Mingsheng Fang, Shiyi Lan, Jan Kautz, and
  Jose~M. Alvarez.
\newblock Fb-occ: 3d occupancy prediction based on forward-backward view
  transformation.
\newblock \emph{arXiv}, 2023{\natexlab{b}}.

\bibitem[Lin et~al.(2017{\natexlab{a}})Lin, Doll{\'a}r, Girshick, He,
  Hariharan, and Belongie]{lin2017feature}
Tsung-Yi Lin, Piotr Doll{\'a}r, Ross Girshick, Kaiming He, Bharath Hariharan,
  and Serge Belongie.
\newblock Feature pyramid networks for object detection.
\newblock In \emph{Proceedings of the IEEE conference on computer vision and
  pattern recognition}, pages 2117--2125, 2017{\natexlab{a}}.

\bibitem[Lin et~al.(2017{\natexlab{b}})Lin, Goyal, Girshick, He, and
  Dollár]{8237586}
Tsung-Yi Lin, Priya Goyal, Ross Girshick, Kaiming He, and Piotr Dollár.
\newblock Focal loss for dense object detection.
\newblock In \emph{2017 IEEE International Conference on Computer Vision
  (ICCV)}, pages 2999--3007, 2017{\natexlab{b}}.

\bibitem[Liu et~al.(2024)Liu, Chen, Wang, Yang, Li, Zeng, Chen, Li, and
  Wang]{Liu2023}
Haisong Liu, Yang Chen, Haiguang Wang, Zetong Yang, Tianyu Li, Jia Zeng, Li
  Chen, Hongyang Li, and Limin Wang.
\newblock Fully sparse 3d occupancy prediction.
\newblock In \emph{European Conference on Computer Vision}, pages 54--71.
  Springer, 2024.

\bibitem[Loshchilov and Hutter(2019)]{loshchilov2018decoupled}
Ilya Loshchilov and Frank Hutter.
\newblock Decoupled weight decay regularization.
\newblock In \emph{International Conference on Learning Representations}, 2019.

\bibitem[Ma et~al.(2024)Ma, Tan, Qu, Ma, Zhang, and Xie]{Ma2024}
Qihang Ma, Xin Tan, Yanyun Qu, Lizhuang Ma, Zhizhong Zhang, and Yuan Xie.
\newblock Cotr: Compact occupancy transformer for vision-based 3d occupancy
  prediction.
\newblock In \emph{IEEE/CVF Conference on Computer Vision and Pattern
  Recognition (CVPR)}, pages 19936--19945, 2024.

\bibitem[Mao et~al.(2023)Mao, Shi, Wang, and Li]{Mao2023}
Jiageng Mao, Shaoshuai Shi, Xiaogang Wang, and Hongsheng Li.
\newblock 3d object detection for autonomous driving: A comprehensive survey.
\newblock \emph{International Journal of Computer Vision}, 131:\penalty0
  1909--1963, 2023.

\bibitem[Pan et~al.(2023)Pan, Liu, Liu, Huang, Wang, Zhang, Xu, Lai, and
  Yang]{Pan2023}
Mingjie Pan, Li Liu, Jiaming Liu, Peixiang Huang, Longlong Wang, Shanghang
  Zhang, Shaoqing Xu, Zhiyi Lai, and Kuiyuan Yang.
\newblock Uniocc: Unifying vision-centric 3d occupancy prediction with
  geometric and semantic rendering.
\newblock \emph{arXiv}, 2023.

\bibitem[Pan et~al.(2024)Pan, Liu, Zhang, Huang, Li, Xie, Wang, Liu, and
  Zhang]{Pan2024}
Mingjie Pan, Jiaming Liu, Renrui Zhang, Peixiang Huang, Xiaoqi Li, Hongwei Xie,
  Bing Wang, Li Liu, and Shanghang Zhang.
\newblock Renderocc: Vision-centric 3d occupancy prediction with 2d rendering
  supervision.
\newblock \emph{Proceedings - IEEE International Conference on Robotics and
  Automation}, pages 12404--12411, 2024.

\bibitem[Peng et~al.(2023)Peng, Alcaide, Anthony, Albalak, Arcadinho, Biderman,
  Cao, Cheng, Chung, Du, Grella, Kiran, He, Hou, Lin, Kazienko, Kocon, Kong,
  Koptyra, Lau, Mantri, Mom, Saito, Song, Tang, Wang, Wind, Woźniak, Zhang,
  Zhang, Zhao, Zhou, Zhou, Zhu, and Zhu]{Peng2023}
Bo Peng, Eric Alcaide, Quentin Anthony, Alon Albalak, Samuel Arcadinho, Stella
  Biderman, Huanqi Cao, Xin Cheng, Michael Chung, Xingjian Du, Matteo Grella,
  G.~V.~Kranthi Kiran, Xuzheng He, Haowen Hou, Jiaju Lin, Przemysław Kazienko,
  Jan Kocon, Jiaming Kong, Bartłomiej Koptyra, Hayden Lau, Krishna Sri~Ipsit
  Mantri, Ferdinand Mom, Atsushi Saito, Guangyu Song, Xiangru Tang, Bolun Wang,
  Johan~S. Wind, Stanisław Woźniak, Ruichong Zhang, Zhenyuan Zhang, Qihang
  Zhao, Peng Zhou, Qinghua Zhou, Jian Zhu, and Rui~Jie Zhu.
\newblock Rwkv: Reinventing rnns for the transformer era.
\newblock In \emph{Conference on Empirical Methods in Natural Language
  Processing}, pages 14048--14077. Association for Computational Linguistics
  (ACL), 2023.

\bibitem[Sirohi et~al.(2021)Sirohi, Mohan, B{\"u}scher, Burgard, and
  Valada]{sirohi2021efficientlps}
Kshitij Sirohi, Rohit Mohan, Daniel B{\"u}scher, Wolfram Burgard, and Abhinav
  Valada.
\newblock Efficientlps: Efficient lidar panoptic segmentation.
\newblock \emph{IEEE Transactions on Robotics}, 38\penalty0 (3):\penalty0
  1894--1914, 2021.

\bibitem[Sze and Kunze(2024)]{Sze2024}
Samuel Sze and Lars Kunze.
\newblock Real-time 3d semantic occupancy prediction for autonomous vehicles
  using memory-efficient sparse convolution.
\newblock In \emph{IEEE Intelligent Vehicles Symposium}, pages 1286--1293.
  Institute of Electrical and Electronics Engineers Inc., 2024.

\bibitem[Szeliski(2022)]{Szeliski2022}
Richard Szeliski.
\newblock \emph{Computer vision: algorithms and applications}.
\newblock Springer Nature, 2022.

\bibitem[Tian et~al.(2023)Tian, Jiang, Yun, Mao, Yang, Wang, Wang, and
  Zhao]{Tian2023}
Xiaoyu Tian, Tao Jiang, Longfei Yun, Yucheng Mao, Huitong Yang, Yue Wang, Yilun
  Wang, and Hang Zhao.
\newblock Occ3d: A large-scale 3d occupancy prediction benchmark for autonomous
  driving.
\newblock \emph{Advances in Neural Information Processing Systems},
  36:\penalty0 64318--64330, 2023.

\bibitem[Tong et~al.(2023)Tong, Sima, Wang, Chen, Wu, Deng, Gu, Lu, Luo, Lin,
  and Li]{Tong2023}
Wenwen Tong, Chonghao Sima, Tai Wang, Li Chen, Silei Wu, Hanming Deng, Yi Gu,
  Lewei Lu, Ping Luo, Dahua Lin, and Hongyang Li.
\newblock Scene as occupancy.
\newblock In \emph{IEEE/CVF International Conference on Computer Vision
  (ICCV)}, pages 8406--8415, 2023.

\bibitem[Wang et~al.(2024{\natexlab{a}})Wang, Wang, Tang, Zheng, Ren, Feng, and
  Ma]{Wang2024Gen}
Guoqing Wang, Zhongdao Wang, Pin Tang, Jilai Zheng, Xiangxuan Ren, Bailan Feng,
  and Chao Ma.
\newblock Occgen: Generative multi-modal 3d occupancy prediction for autonomous
  driving.
\newblock \emph{arXiv}, 2024{\natexlab{a}}.

\bibitem[Wang et~al.(2024{\natexlab{b}})Wang, Yin, Long, Zhang, Xing, Guo, and
  Zhang]{Wang2024}
Junming Wang, Wei Yin, Xiaoxiao Long, Xingyu Zhang, Zebin Xing, Xiaoyang Guo,
  and Qian Zhang.
\newblock Occrwkv: Rethinking efficient 3d semantic occupancy prediction with
  linear complexity.
\newblock \emph{arXiv}, 2024{\natexlab{b}}.

\bibitem[Wang et~al.(2024{\natexlab{c}})Wang, Wang, Yu, Li, Feng, Chen, and
  Zhu]{Wang2024Reli}
Song Wang, Zhongdao Wang, Jiawei Yu, Wentong Li, Bailan Feng, Junbo Chen, and
  Jianke Zhu.
\newblock Reliocc: Towards reliable semantic occupancy prediction via
  uncertainty learning.
\newblock \emph{arXiv}, 2024{\natexlab{c}}.

\bibitem[Wang et~al.(2023)Wang, Zhu, Xu, Zhang, Wei, Chi, Ye, Du, Lu, and
  Wang]{Wang2023}
Xiaofeng Wang, Zheng Zhu, Wenbo Xu, Yunpeng Zhang, Yi Wei, Xu Chi, Yun Ye,
  Dalong Du, Jiwen Lu, and Xingang Wang.
\newblock Openoccupancy: A large scale benchmark for surrounding semantic
  occupancy perception.
\newblock In \emph{IEEE/CVF International Conference on Computer Vision
  (ICCV)}, pages 17850--17859, 2023.

\bibitem[Wang et~al.(2024{\natexlab{d}})Wang, Chen, Liao, Fan, and
  Zhang]{Wang2024PanoOcc}
Yuqi Wang, Yuntao Chen, Xingyu Liao, Lue Fan, and Zhaoxiang Zhang.
\newblock Panoocc: Unified occupancy representation for camera-based 3d
  panoptic segmentation.
\newblock In \emph{IEEE/CVF Conference on Computer Vision and Pattern
  Recognition (CVPR)}, pages 17158--17168, 2024{\natexlab{d}}.

\bibitem[Wei et~al.(2023)Wei, Zhao, Zheng, Zhu, Zhou, and Lu]{Wei2023}
Yi Wei, Linqing Zhao, Wenzhao Zheng, Zheng Zhu, Jie Zhou, and Jiwen Lu.
\newblock Surroundocc: Multi-camera 3d occupancy prediction for autonomous
  driving.
\newblock In \emph{IEEE/CVF International Conference on Computer Vision
  (ICCV)}, pages 21729--21740, 2023.

\bibitem[Xu et~al.(2023)Xu, Liu, Ning, Zhao, Cheng, and Nie]{Xu2023}
Guixing Xu, Wei Liu, Zuotao Ning, Qixi Zhao, Shuai Cheng, and Jiwei Nie.
\newblock 3d semantic scene completion and occupancy prediction for autonomous
  driving: A survey.
\newblock \emph{2023 4th International Conference on Computers and Artificial
  Intelligence Technology, CAIT 2023}, pages 181--188, 2023.

\bibitem[Ye et~al.(2022)Ye, Zhou, Chen, Xie, Wang, Wang, and
  Foroosh]{ye2022lidarmultinet}
Dongqiangzi Ye, Zixiang Zhou, Weijia Chen, Yufei Xie, Yu Wang, Panqu Wang, and
  Hassan Foroosh.
\newblock Lidarmultinet: Towards a unified multi-task network for lidar
  perception.
\newblock \emph{arXiv preprint arXiv:2209.09385}, 2022.

\bibitem[Yu et~al.(2023)Yu, Shu, Deng, Lu, Liu, Yu, Yang, Li, and Chen]{Yu2023}
Zichen Yu, Changyong Shu, Jiajun Deng, Kangjie Lu, Zongdai Liu, Jiangyong Yu,
  Dawei Yang, Hui Li, and Yan Chen.
\newblock Flashocc: Fast and memory-efficient occupancy prediction via
  channel-to-height plugin.
\newblock \emph{arXiv}, 2023.

\bibitem[Yu et~al.(2024)Yu, Shu, Sun, Linghu, Wei, Yu, Liu, Yang, Li, and
  Chen]{Yu2024}
Zichen Yu, Changyong Shu, Qianpu Sun, Junjie Linghu, Xiaobao Wei, Jiangyong Yu,
  Zongdai Liu, Dawei Yang, Hui Li, and Yan Chen.
\newblock Panoptic-flashocc: An efficient baseline to marry semantic occupancy
  with panoptic via instance center.
\newblock \emph{arXiv}, 2024.

\bibitem[Zhang et~al.(2023)Zhang, Zhu, and Du]{Zhang2023}
Yunpeng Zhang, Zheng Zhu, and Dalong Du.
\newblock Occformer: Dual-path transformer for vision-based 3d semantic
  occupancy prediction.
\newblock In \emph{IEEE/CVF International Conference on Computer Vision
  (ICCV)}, pages 9433--9443, 2023.

\bibitem[Zhang et~al.(2024)Zhang, Zhang, Wang, Xu, and Huang]{Zhang2024}
Yanan Zhang, Jinqing Zhang, Zengran Wang, Junhao Xu, and Di Huang.
\newblock Vision-based 3d occupancy prediction in autonomous driving: a review
  and outlook.
\newblock \emph{arXiv}, 2024.

\bibitem[Zhao et~al.(2024)Zhao, Xu, Wang, Zhang, Zhang, Zheng, Du, Zhou, and
  Lu]{Zhao2024}
Linqing Zhao, Xiuwei Xu, Ziwei Wang, Yunpeng Zhang, Borui Zhang, Wenzhao Zheng,
  Dalong Du, Jie Zhou, and Jiwen Lu.
\newblock Lowrankocc: Tensor decomposition and low-rank recovery for
  vision-based 3d semantic occupancy prediction.
\newblock In \emph{IEEE/CVF Conference on Computer Vision and Pattern
  Recognition (CVPR)}, pages 9806--9815, 2024.

\bibitem[Zheng et~al.(2024)Zheng, Li, Li, Zheng, Jin, Zhong, Long, Zhao, and
  Zhang]{Zheng2024}
Yupeng Zheng, Xiang Li, Pengfei Li, Yuhang Zheng, Bu Jin, Chengliang Zhong,
  Xiaoxiao Long, Hao Zhao, and Qichao Zhang.
\newblock Monoocc: Digging into monocular semantic occupancy prediction.
\newblock \emph{Proceedings - IEEE International Conference on Robotics and
  Automation}, pages 18398--18405, 2024.

\bibitem[Zhou et~al.(2021)Zhou, Zhang, and Foroosh]{zhou2021panoptic}
Zixiang Zhou, Yang Zhang, and Hassan Foroosh.
\newblock Panoptic-polarnet: Proposal-free lidar point cloud panoptic
  segmentation.
\newblock In \emph{Proceedings of the IEEE/CVF Conference on Computer Vision
  and Pattern Recognition}, pages 13194--13203, 2021.

\bibitem[Zhu et~al.(2020)Zhu, Su, Lu, Li, Wang, and Dai]{zhu2020deformable}
Xizhou Zhu, Weijie Su, Lewei Lu, Bin Li, Xiaogang Wang, and Jifeng Dai.
\newblock Deformable detr: Deformable transformers for end-to-end object
  detection.
\newblock In \emph{International Conference on Learning Representations}, 2020.

\end{thebibliography}
